\documentclass{article}

\usepackage{amsmath, amssymb, amsthm}
\usepackage[sort&compress, numbers]{natbib} 
\usepackage{graphicx}

\newtheorem{theorem}{Theorem}
\newtheorem{corollary}{Corollary}
\newtheorem{lemma}{Lemma}
\newtheorem{proposition}{Proposition}
\newtheorem{remark}{Remark}
\newtheorem{definition}{Definition}
\theoremstyle{boldassumption}

\usepackage[a4paper, total={6in, 8in}]{geometry}

\usepackage{color}
\definecolor{darkred}{RGB}{100,0,0}
\definecolor{darkgreen}{RGB}{0,100,0}
\definecolor{darkblue}{RGB}{0,0,150}

\definecolor{purple}{rgb}{0.4,.1,.9}

\newenvironment{myproof}[1][]{%
    \par\addvspace{\topsep}%
    \noindent\textbf{Proof #1.}\ \ignorespaces%
}{%
    \hfill$\blacksquare$\par\addvspace{\topsep}%
}

\title{Stable Density Ridges: Consistency and Convergence of Subspace Constrained Mean Shift}
\author{Wanli Qiao}
\date{\today}

\begin{document}

\maketitle

\begin{abstract}
The Subspace Constrained Mean Shift (SCMS) algorithm is a popular nonparametric method for extracting density ridges, which serve as a low-dimensional representation of high-dimensional data. It is a widely held belief in the literature that SCMS trajectories converge to the classical density ridge, which we call the ``static ridge'', defined via the density gradient and the eigenvalues and eigenvectors of the density's Hessian. In this paper, we demonstrate that this assumption does not hold in general, as the static definition fails to account for the rotation of the trailing eigenspace along the continuous flow of the algorithm's underlying vector field. To resolve this, we propose a paradigm shift by introducing the ``stable ridge'', a novel geometric structure defined through the lens of dynamical systems and the Jacobian of the projected density gradient. We prove that this stable ridge is the true theoretical target of the SCMS algorithm. Building upon this foundation, we develop a generalized SCMS framework utilizing a constant step size, establishing its uniform R-linear convergence and topological surjectivity onto the stable ridge. We further derive the rates of convergence for estimating the stable ridge in terms of the Hausdorff distance. Finally, we expose that the original SCMS algorithm suffers from polynomial-time computational complexity, which is caused by implicitly coupling the step size to the smoothing bandwidth via the Mean Shift operator, and demonstrate how our generalized framework provides a statistically consistent and more efficient solution.

\vspace{1em}
\end{abstract}

\section{Introduction}
\label{sec:introduction}

The extraction of low-dimensional geometric structures from high-dimensional data is a fundamental problem in modern statistics and machine learning. In many applications, data naturally concentrate around lower-dimensional manifolds, such as curves and surfaces, driving the need for methods of non-linear dimension reduction \citep{tenenbaum2000global, roweis2000nonlinear} and manifold fitting~\citep{fefferman2018fitting,yao2025manifold,yao2023manifold}. Early approaches to modeling these structures, such as principal curves and surfaces \citep{hastie1989principal}, provided frameworks by seeking manifolds that pass through the middle of a data cloud to minimize projection error. While subsequent advancements theoretically guaranteed the existence of these curves by imposing global regularization constraints, such as bounding the total length \citep{kegl2000learning} or curvature \citep{sandilya2002principal}, these formulations remain tied to global optimization heuristics. Because they rely on minimizing projection errors subject to rigid structural penalties, they often struggle to flexibly adapt to variations in local data density.

As a compelling alternative, the concept of the \textit{density ridge} emerged to define the hidden skeleton of a dataset directly through the local differential geometry of the underlying probability density function. By tracking the regions of highest density constrained by local subspaces, density ridges form low-dimensional structures representing the backbone of the data. Rather than imposing arbitrary global length or curvature constraints, they capture the local geometry determined by the data's distribution. In practice, the Subspace Constrained Mean Shift (SCMS) algorithm, developed by \cite{ozertem2011locally}, was designed to extract these ridges from data. Prominent applications include the identification of cosmic web filaments and large-scale structures in astrophysics \citep{chen2015cosmic}, the extraction of seismic fault lines from earthquake data \citep{li2020posterior}, the tracing of anatomical structures such as blood vessels in medical imaging \citep{eberly1996ridges}, and the optimization of geospatial police patrol routes by mapping crime hot-spot filaments \citep{moews2021filaments}. In this paper, we study the theoretical properties of the SCMS algorithm by clarifying the definition of the ridges it targets, and we quantify its performance in terms of both statistical consistency and computational complexity.

\subsection{The Static Ridge and the SCMS Algorithm}

We first introduce the concept of density ridges used in the literature. Consider a probability density function $f \in C^2(\mathbb{R}^d, \mathbb{R})$ for $d\ge2$. Let $k \in \{1, \dots, d-1\}$ be a fixed integer representing the dimension of the target manifold. For any point $x \in \mathbb{R}^d$, let $\lambda_1(x) \ge \dots \ge \lambda_d(x)$ be the sorted eigenvalues of the Hessian $\nabla^2 f(x)$. Let $E_\parallel(x)$ denote the $k$-dimensional subspace spanned by the eigenvectors corresponding to the top $k$ eigenvalues of $\nabla^2 f(x)$, and let $E_\perp(x)$ denote the $(d-k)$-dimensional subspace spanned by the trailing $d-k$ eigenvectors. We refer to $E_\parallel(x)$ as the leading eigenspace and $E_\perp(x)$ as the trailing eigenspace of the density Hessian. Let $V(x)$ be the $d \times (d-k)$ matrix whose columns form an orthonormal basis for the trailing eigenspace $E_\perp(x)$. We define the orthogonal projection matrix onto this trailing eigenspace as $\Pi(x) = V(x)V(x)^\top$, and the projected density gradient onto this trailing eigenspace is denoted $\xi(x) = \Pi(x)\nabla f(x)$.

The ridge concept uses the gradient and Hessian (and its eigenvalues and eigenvectors) of the density, and traces its origins to the image processing and computer vision literature \citep{eberly1996ridges}. We will refer to such a ridge concept as the \textit{static ridge}, defined as
\begin{equation}\label{eq:static_ridge}
\mathcal{R}_{\text{static}}(f) = \left\{ x \in \mathbb{R}^d :\; \xi(x) = 0 \text{ and } \lambda_{k+1}(x) < 0 \right\}.
\end{equation}
While this is often referred to in the literature as $k$-dimensional ridges (or $k$-ridges), we treat $k$ as a fixed parameter throughout this paper and omit it from our notation for brevity. 
A point belongs to the static ridge if the density gradient is orthogonal to the trailing eigenspace, where the density is a local maximum along any direction in the trailing eigenspace. This geometric interpretation can be seen from the fact that the eigenvalues represent the second directional derivatives of the density along fixed eigenvector directions. Hence the static ridge extends the concept of a local maximum to a set of local maxima constrained in particular subspaces, which forms a $k$-dimensional structure.

In practice, the population density $f$ is unknown and should be estimated from a finite sample $X_1, \dots, X_n \sim f$. This is typically achieved using a Kernel Density Estimator (KDE) with a smoothing kernel $K$ and bandwidth $h > 0$. In the statistical literature, there are a series of papers on the statistical inference for the static ridge using the plug-in estimator based on KDE \citep{genovese2014nonparametric,chen2015asymptotic,qiao2016theoretical,qiao2021asymptotic,qiao2025confidence}. 

To extract the static ridge from the estimated density, the SCMS algorithm developed in \cite{ozertem2011locally} also utilizes the geometric intuition above for the static ridge. The algorithm builds upon the seminal Mean Shift algorithm, which is a nonparametric mode-seeking technique widely used in clustering and computer vision \citep{fukunaga1975estimation, cheng1995mean, comaniciu2002mean}. The Mean Shift operator, denoted ${\sf MS}_{n,h}: \mathbb{R}^d \to \mathbb{R}^d$, at each point $x$ computes a vector pointing toward the locally weighted mass center using neighboring data points of $x$. The statistical properties of Mean Shift algorithm and relevant hill-climbing algorithms for clustering have been studied in \cite{chacon2015population,arias2016errata,chen2017statistical,qiao2022space,arias2023moving,arias2023unifying,arias2025clustering}. 

\begin{figure}[htbp]
    \centering
    \includegraphics[width=0.7\textwidth]{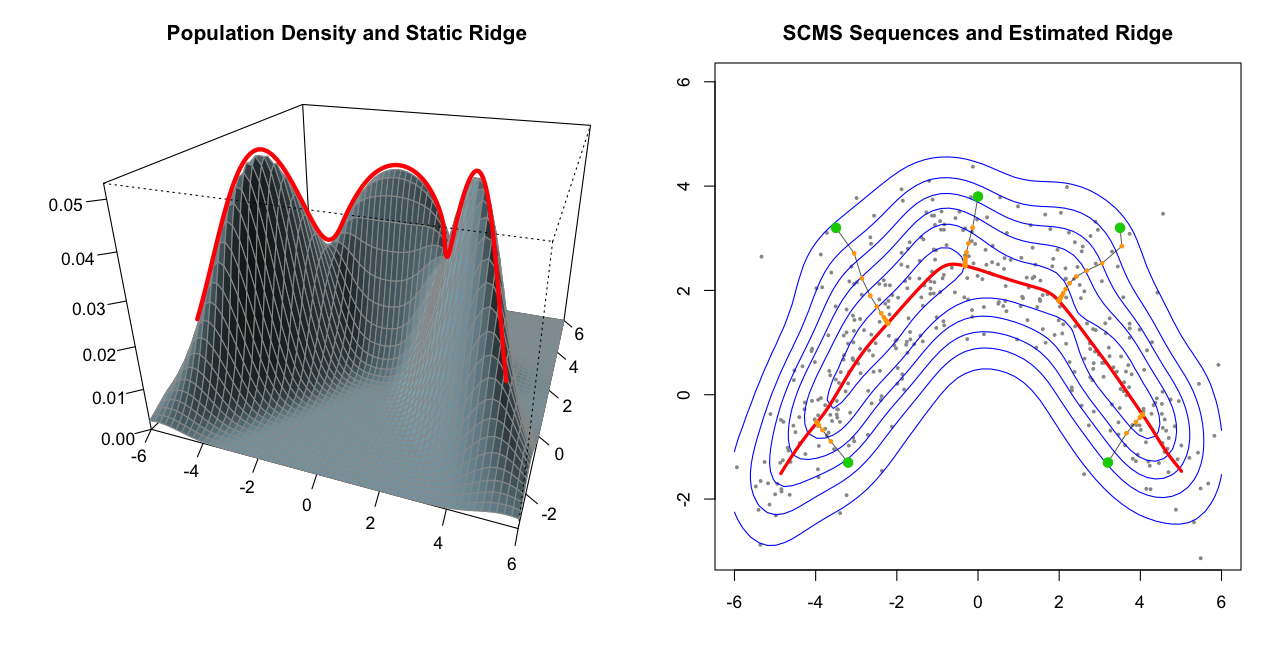}
   \caption{Visualization of the static ridge and SCMS. 
Left panel: A 3D perspective plot of the population density $f$, modeled as a 3-component Gaussian mixture, where the solid red line represents the static ridge. 
Right panel: SCMS trajectories. The grey dots are data samples drawn from the population model, and the blue contours represent the resulting KDE. Green dots indicate the starting points for the SCMS algorithm. The orange points illustrate the path of the algorithm. The trajectories are seen to converge onto the estimated ridge of the log-KDE (in red).}
    \label{fig:ridge_illustration}
\end{figure}

In particular, \cite{ozertem2011locally} utilize a Gaussian kernel, under which the Mean Shift vector becomes proportional to the gradient of the estimated log-density, that is, ${\sf MS}_{n,h}(x) \propto \nabla \log \widehat{f}(x)$. The SCMS algorithm iteratively updates a test point $x_m$ by projecting this Mean Shift vector onto an estimated trailing eigenspace. In their original formulation, this trailing eigenspace is determined by the eigenvectors of the \textit{log-density} Hessian, $\nabla^2 \log \widehat{f}(x)$. Denoting the corresponding projection matrix as $\widehat{\Pi}^{\log}(x)$, the update rule for the SCMS algorithm in \cite{ozertem2011locally} is
\begin{equation}\label{eq:scms_discrete_update}
x_{m+1} = x_m + \widehat{\Pi}^{\log}(x_m) {\sf MS}_{n,h}(x_m).
\end{equation}
See Figure~\ref{fig:ridge_illustration} for an illustration of the SCMS algorithm's performance in ridge estimation.

Because both the Mean Shift operator and the projection matrix are based on $\log \widehat{f}(x)$, the SCMS algorithm aims for the ridge based on the log-density rather than the raw density as the target. Despite this distinction, it became a widely held belief in the literature \citep{ghassabeh2013some,genovese2014nonparametric,zhang2023linear,zhang2026mode} that the trajectories of SCMS converged to the static ridge $\mathcal{R}_{\text{static}}(f)$ (or its log-density equivalent) in the asymptotic setting when $n\to\infty$ and $m\to\infty$.

However, a recent discovery by \cite{qiao2025algorithms} demonstrated that the relationship between SCMS and the static ridge is more complex than previously assumed. As illustrated by their counterexample (also presented in Appendix \ref{Pitchfork}), the SCMS algorithm does not necessarily converge to the static ridge. The intuition behind this divergence lies in the nature of the SCMS vector field: as a point flows along the trajectory of $\xi$, the trailing eigenspace $\Pi$ itself dynamically rotates. Because the static ridge is defined by pointwise directional derivatives, it fails to account for the continuous rotation of the projection matrix along a flow line. To resolve this, \cite{qiao2025algorithms} developed new algorithms that have a statistical guarantee to converge to the static ridge $\mathcal{R}_{\text{static}}(f)$. 

This still leaves a critical open problem: \emph{if the SCMS algorithm does not estimate the static ridge, what structure is it actually finding?}

\subsection{The Stable Ridge: From Geometry to Dynamics}

In this paper, we propose a paradigm shift by introducing the concept of the \textit{stable ridge}, and proving that it is this structure that the SCMS algorithm actually targets and extracts.

Unlike the static ridge, which is defined by the Hessian $\nabla^2 f(x)$, the stable ridge is defined by the topology of the dynamical system governed by $\xi$. To ensure the stability of this system is well-defined, we assume the density $f \in C^3(\mathbb{R}^d, \mathbb{R})$. This higher-order smoothness is necessary because the Jacobian $J_\xi := \nabla \xi$ incorporates the derivatives of the projection matrix $\Pi$. For any $x\in\mathbb{R}^d$, let $\mu_1(x), \dots, \mu_d(x)$ denote the eigenvalues of this (asymmetric) Jacobian sorted by their real parts in descending order, that is, $\mathrm{Re}(\mu_1(x)) \ge \dots \ge \mathrm{Re}(\mu_d(x))$, where for a complex number $z$,  $\mathrm{Re}(z)$ denotes its real part. We define the stable ridge as
\begin{equation}\label{eq:stable_ridge}
\mathcal{R}_{\text{stable}}(f) = \left\{ x \in \mathbb{R}^d :\; \xi(x) = 0 \text{ and } \mathrm{Re}(\mu_{k+1}(x)) < 0 \right\}.
\end{equation}

Compared to the static ridge definition in \eqref{eq:static_ridge}, the stable ridge differs only by replacing the condition $\lambda_{k+1}(x) < 0$ with $\mathrm{Re}(\mu_{k+1}(x)) < 0$. This subtle spectral shift, however, reflects a fundamental conceptual transition from a static geometric perspective to a dynamical systems approach. In classical optimization, local attractors are defined by the zeros of the gradient vector field (i.e., $\nabla f(x) = 0$), with stability governed by the negative eigenvalues of its Jacobian, which is the Hessian $\nabla^2 f(x)$. Our dynamic framework mirrors this structure for the vector field defined by the \textit{projected  gradient} $\xi$: the stable ridge is defined by $\xi(x)=0$, with stability governed by the negative real parts of the eigenvalues of its Jacobian, $J_\xi(x)$. This extension can be clearly illustrated by examining the extreme case where the dimension is $k = 0$. Under the static definition \eqref{eq:static_ridge}, setting $k=0$ requires all $d$ eigenvalues of the Hessian to be negative ($\lambda_1(x) < 0$), which recovers the standard second-order condition for a local maximum (or mode). Under our stable definition \eqref{eq:stable_ridge}, $k=0$ implies the trailing eigenspace spans the entirety of $\mathbb{R}^d$, such that the projection matrix $\Pi(x)$ becomes the identity matrix $I_d$, the projected gradient $\xi(x)$ reduces to the raw gradient $\nabla f(x)$, and $J_\xi(x)$ is simply the Hessian $\nabla^2 f(x)$. Thus, the stable ridge also collapses to the set of local maxima. However, this comparison highlights the fundamental shift in perspective: while the static definition views the mode simply as a peak, the stable definition characterizes it as an equilibrium point that acts as an attractor from all $d$ directions in the dynamical system.

However, when the dimension $k \ge 1$, these two concepts of static and stable ridges are generally not equivalent. Because the SCMS algorithm can be viewed as a discretization of a continuous dynamical system, the stable ridge---rather than the static ridge---serves as the correct theoretical target for its convergence, the rigorous proof of which is the overarching goal of this paper. A detailed mathematical derivation quantifying the divergence and coincidence between the static and stable ridge definitions (specified when $d=2$ and $k=1$) with illustrative examples, is provided in Appendix \ref{app:topological_divergence}.

\textbf{Paper Organization.} The remainder of this paper is organized as follows. Section \ref{sec:preliminaries} introduces the ridge-regular function class as part of our core assumptions for the model. Section \ref{sec:population_geometry} characterizes the geometric properties of the stable ridge, studies the continuous flow driven by the vector field $\xi$, and shows its capability of surjective recovery of the full stable ridge. Section \ref{sec:discrete_optimization} transitions to the discrete algorithmic setting, where we develop the generalized SCMS algorithm (still based on the true density) using a constant step size, and establish its uniform R-linear convergence and topological surjectivity. Section \ref{sec:finite_sample_consistency} introduces the sample version of the generalized SCMS based on KDE and derives the convergence rate of the Hausdorff error in estimating the true stable ridge. Section \ref{sec:original_scms_log_density} bridges our generalized framework back to the original SCMS algorithm operating on the log-density. There, we expose the polynomial-time computational complexity of the original SCMS, which is caused by implicitly coupling the step size to the smoothing bandwidth via the Mean Shift operator, and demonstrate how our generalized framework with a constant step size provides a more efficient solution. Section \ref{sec:numerical_experiments} provides numerical results validating both the statistical consistency and the computational complexities derived in the text. Finally, Appendix \ref{app:topological_divergence} provides examples demonstrating the relationship between the static and stable ridge definitions, and Appendix \ref{app:proofs} contains the proofs for all the theoretical results in this paper.

\section{Assumptions and the Ridge-Regular Class}
\label{sec:preliminaries}

Let $\mathcal{D} \subset \mathbb{R}^d$ be a compact and connected domain of interest for ridge extraction, so that 
we focus our analysis on the restricted stable ridge $\mathcal{R}(f)$ defined as
\begin{equation}\label{eq:restricted_stable_ridge}
\mathcal{R}(f) = \mathcal{R}_{\text{stable}}(f) \cap \mathcal{D} = \left\{ x \in \mathcal{D} :\; \xi(x) = 0 \text{ and } \mathrm{Re}(\mu_{k+1}(x)) < 0 \right\}.
\end{equation}

We introduce some regularity assumptions used in this paper.

\begin{description}
    \item[Assumption (A1).] 
The density function satisfies $f \in C^4(\mathbb{R}^d, \mathbb{R})$ with its partial derivatives up to the fourth order uniformly bounded over $\mathbb{R}^d$. Furthermore, there exists a constant $c_0 > 0$ such that $f(x) \ge c_0$ for all $x \in \mathcal{D}$.
\end{description}

While $C^3$ smoothness is sufficient to define the stable ridge, we require $C^4$ smoothness to study the consistency and convergence of the SCMS algorithms. Note that while smoothness is formulated on $\mathbb{R}^d$ to accommodate infinite-support kernels such as the Gaussian kernel, if we choose a kernel with compact support, then this assumption only needs to hold within a neighborhood of $\mathcal{D}$.

To define the geometric conditions required for stable ridge extraction, we collect the requirements into a function class. 

For any function $g \in C^3(\mathbb{R}^d, \mathbb{R})$, let the quantities, sets and operators $\lambda_j^g(x)$, $E_\perp^g(x)$, $\Pi_g(x)$, $\xi_g(x)$, and $\mu_j^g(x)$ be defined analogously to those of $f$. For any $\epsilon\ge0$, define $$\mathcal{R}_{\epsilon}(g) = \{x \in \mathcal{D} : \|\xi_g(x)\| \le \epsilon \text{ and } \mathrm{Re}(\mu_{k+1}^g(x)) < 0\},$$
such that $\mathcal{R}_{0}(f)=\mathcal{R}(f)$.

\begin{definition}
\label{def:function_class}
The function $g$ is said to belong to the ridge-regular class $\mathcal{F}(\eta, \gamma, \kappa, \mathcal{D})$ for some constants $\eta,\gamma,\kappa>0$, if it satisfies the following three properties for all $x \in \mathcal{D}$.
\begin{enumerate}
    \item $\lambda_k^g(x) - \lambda_{k+1}^g(x) \ge \eta$.
    \item The set $\mathcal{R}_{\kappa}(g)$ is contained within the interior of $\mathcal{D}$ denoted by $\mathrm{int}(\mathcal{D})$, and 
    \begin{equation*}
    \sup \{ u^\top J_{\xi_g}(x) u:  u \in E_\perp^g(x), \; \|u\| = 1\}\le -\gamma.
    \end{equation*}
    \item $ \inf \left\{ \|\xi_g(x)\| :\; x \in \mathcal{D}, \; \mathrm{Re}(\mu_{k+1}^g(x)) \ge -\frac{\gamma}{2} \right\} \ge \kappa.$
\end{enumerate}
\end{definition}

We assume the population density satisfies these geometric constraints.

\begin{description}
    \item[Assumption (A2).] 
$f \in \mathcal{F}(\eta, \gamma, \kappa, \mathcal{D})$ for some positive constants $\eta, \gamma$, and $\kappa$.
\end{description}

\begin{remark}
\label{rem:function_class_implications}
The properties required for $f\in \mathcal{F}(\eta, \gamma, \kappa, \mathcal{D})$ guarantee the statistical and computational stability of the SCMS algorithms. Property 1 provides the necessary spectral gap to ensure the projection matrix $\Pi$ is uniquely defined and smooth on $\mathcal{D}$. Property 2 forces the symmetric part of the Jacobian to be negative definite when restricted to the trailing eigenspace; this guarantees the monotonic decay of $\|\xi\|$ along the SCMS trajectories. Finally, Property 3 regularizes the vector field $\xi$ far away from the ridge to prevent the emergence of spurious features under small perturbations.
\end{remark}

\textbf{Notation and Global Constants.} To streamline the theoretical analysis in the subsequent sections, we define the following global constants.
\begin{equation*}
L = \sup_{x \in \mathcal{D}} \|J_\xi(x)\| \quad \text{and} \quad M = \frac{1}{2} \sup_{x \in \mathcal{D}} \|\nabla^2 \xi(x)\|,
\end{equation*}
where for a matrix $A$, $\|A\|$ denotes its induced spectral norm; for a vector field $v: \mathbb{R}^d \to \mathbb{R}^d$, its Hessian $\nabla^2 v(x)$ is a bilinear map, and we define its multilinear operator norm as $\|\nabla^2 v(x)\| = \sup_{\|u_1\|=1, \|u_2\|=1} \|\nabla^2 v(x)[u_1, u_2]\|$. Based on Assumptions (A1) and (A2), it is straightforward to see $L<\infty$ and $M<\infty$. Furthermore, we define $$d_{\partial} = \inf \left\{ \|x - y\| : x \in \mathcal{R}_{\kappa}(f), y \in \partial \mathcal{D} \right\}.$$ It can be shown that $\mathcal{R}_{\kappa}(f)$ is a compact set (see the proof of Lemma \ref{lem:geometric_regularity} below). Hence $d_{\partial}$ is positive based on Property 2 of $\mathcal{F}(\eta, \gamma, \kappa, \mathcal{D})$.

We collect some notation used throughout the paper. Let $A$ and $B$ be two non-empty subsets of $\mathbb{R}^d$. The Hausdorff distance between $A$ and $B$ is
\begin{equation}
    d_H(A, B) = \max \left\{d(A | B) , \;  d(B|A)\right\},
\end{equation}
where $d(A | B) = \sup_{x \in A} \inf_{y \in B} \|x - y\|$ and $d(B|A) = \sup_{y \in B} \inf_{x \in A} \|x - y\|$. For any $x \in \mathbb{R}^d$ and $r > 0$, let $B_r(x)$ denote the closed Euclidean ball of radius $r$ centered at $x$, that is, $B_r(x) = \{y \in \mathbb{R}^d : \|y - x\| \le r\}$. Finally, for positive sequences $a_n$ and $b_n$, we write $a_n = O(b_n)$ if there exists a constant $C > 0$ such that $a_n \le C b_n$ for all $n$ large enough, and we write $a_n \asymp b_n$ if both $a_n = O(b_n)$ and $b_n = O(a_n)$ hold.

\section{Geometry of the Stable Ridge and Continuous SCMS}
\label{sec:population_geometry}

In this section, we describe the geometry of the true stable ridge $\mathcal{R}(f)$ and analyze the behavior of the continuous flow determined by the vector field $\xi$. This continuous flow serves as the infinitesimal limit of the SCMS algorithm at the population level. By understanding the properties of this dynamical system, we lay the groundwork for analyzing the discretization of the flow and the algorithm's convergence and statistical consistency in subsequent sections.

Before tracking the flow, we first confirm that the stable ridge as our target structure is a geometrically well-behaved object under the ridge-regular assumptions.

\begin{lemma}
\label{lem:geometric_regularity}
Under Assumptions (A1) and (A2), the stable ridge $\mathcal{R}(f)$ is a compact, $C^2$-smooth, $k$-dimensional submanifold of $\mathbb{R}^d$ without boundary.
\end{lemma}

With the regularity of the stable ridge established, we turn our attention to the trajectories of the continuous population version of SCMS. Let $\varphi_t(x)$ denote the flow driven by the vector field $\xi$, with the initial value $x$, that is,
\begin{equation}
\label{varphi_def}
\dot{\varphi}_t(x) := \frac{d}{dt}\varphi_t(x) = \xi(\varphi_t(x)), \quad \text{with } \varphi_0(x) = x.
\end{equation}
We define its total arc length as
\begin{equation}
\mathcal{L}(x) = \int_0^\infty \|\xi(\varphi_t(x))\| \, dt.
\end{equation}

As a practical ridge estimation technique, it is crucial for the paths of continuous SCMS initialized near the target ridge remain nearby and converge to the ridge within a finite distance. The following lemma demonstrates that the geometric conditions of the ridge-regular class guarantee an exponential decay in the magnitude of the vector field along the paths, ensuring this convergent behavior.

\begin{lemma}
\label{lem:continuous_limit}
Suppose Assumptions (A1) and (A2) hold. For $\epsilon \in (0, \kappa]$ and $x \in \mathcal{R}_{\epsilon}(f)$, the flow $\varphi_t$ satisfies
\begin{equation}
    \|\xi(\varphi_t(x))\| \le e^{-\gamma t} \|\xi(x)\| \le \|\xi(x)\| \le \epsilon \quad \text{for all } t \ge 0.
\end{equation}
Consequently, $\varphi_t(x) \in \mathcal{R}_{\epsilon}(f)$ for all $t \ge 0$ and
\begin{equation}
    \mathcal{L}(x) \le \frac{1}{\gamma} \|\xi(x)\|.
\end{equation}
Furthermore, $\varphi_t(x)$ converges to a limit point in $\mathcal{R}(f)$ as $t \to \infty$, which we denote $\Phi(x)$, i.e., 
\begin{equation}
\label{Phi def}
\Phi(x) = \lim_{t \to \infty} \varphi_t(x), \text{ for } x\in \mathcal{R}_{\epsilon}(f).
\end{equation}
\end{lemma}

Because any point $x \in \mathcal{R}_{\epsilon}(f)$ flows onto the ridge $\mathcal{R}(f)$, it is informative to quantify how tightly the neighborhood $\mathcal{R}_{\epsilon}(f)$ bounds the true ridge. The following corollary establishes that their Hausdorff distance shrinks linearly with $\epsilon$. The corollary will be useful in quantifying the perturbation of the stable ridge, as a slightly perturbed ridge remains in $ \mathcal{R}_{\epsilon}(f)$ for some small $\epsilon>0$.

\begin{corollary}
\label{cor:hausdorff_bound}
Suppose Assumptions (A1) and (A2) hold. Let $\epsilon$ satisfy $0 \le \epsilon \le \min\big(\kappa, \frac{L\gamma}{M}, \frac{\gamma}{L}\kappa\big)$. Then we have
\begin{equation}
\label{hausdorff_bound}
    \frac{1}{2L} \epsilon \le d_H(\mathcal{R}_\epsilon(f), \mathcal{R}(f)) \le \frac{1}{\gamma} \epsilon.
\end{equation}
\end{corollary}

While Lemma \ref{lem:continuous_limit} ensures that every point in the neighborhood converges to the ridge, it does not guarantee that the \textit{entire} ridge is covered by these limit points. To ensure our extraction process discovers the complete manifold, we need to prove the surjectivity of limit map $\Phi$ defined by the flow $\varphi_t$.

For $\epsilon \in (0, \kappa)$, we define the boundary of the neighborhood as
\begin{equation}
    \partial \mathcal{R}_{\epsilon}(f) = \{x \in \mathcal{D} : \|\xi(x)\| = \epsilon \text{ and } \mathrm{Re}(\mu_{k+1}(x)) < 0\}.
\end{equation}
The following theorem establishes that initializing the flow $\varphi_t$ from $\partial \mathcal{R}_{\epsilon}(f)$ is sufficient to recover every point on the target ridge $\mathcal{R}(f)$.

\begin{theorem}
\label{thm:continuous_surjectivity}
Suppose Assumptions (A1) and (A2) hold. For $\epsilon \in (0, \kappa)$, the limit map $\Phi$ defined in \eqref{Phi def} is surjective from $\partial \mathcal{R}_{\epsilon}(f)$ onto $\mathcal{R}(f)$. That is, for every point $x_* \in \mathcal{R}(f)$, there exists at least one starting point $x_0 \in \partial \mathcal{R}_{\epsilon}(f)$ such that $\Phi(x_0) = x_*$.
\end{theorem}

Finally, for the limit map $\Phi(x)$ to be useful in bounding the errors in subsequent discrete and finite-sample settings, it must be stable under small perturbations. We conclude this section by showing the the continuous differentiability of $\Phi$.

\begin{lemma}
\label{lem:lipschitz_projection}
Suppose Assumptions (A1) and (A2) hold. The limit map $\Phi$ defined in \eqref{Phi def} is continuously differentiable on $\mathcal{R}_{\kappa}(f)$. Furthermore, there exists a uniform constant $L_\Phi > 0$ such that for all $x \in \mathcal{R}_{\kappa}(f)$, we have
\begin{equation}
\|J_\Phi(x)\| \le L_\Phi,
\end{equation}
where $J_\Phi(x)$ is the Jacobian of $\Phi$.
\end{lemma}

\section{Stability and Convergence of the Generalized SCMS Algorithm}
\label{sec:discrete_optimization}

To bridge the gap between the continuous population version of SCMS studied in the previous section and the practical implementation of the algorithm, we analyze the discretized approximation of the flow $\varphi_t$. To distinguish our mathematical framework from the original formulation by \cite{ozertem2011locally}, which couples the step size to the smoothing bandwidth in the algorithm and is implicitly based on the log-density, we define the \textit{generalized SCMS algorithm}. This generalized version operates directly on the vector field $\xi(x)$ derived from the density using a constant step size $\alpha > 0$ specified by the user. 

We define the one-step operator $G_\alpha: \mathcal{D} \to \mathbb{R}^d$ for the discrete algorithm as
\begin{equation}\label{eq:discrete_operator}
G_\alpha(x) = x + \alpha \xi(x).
\end{equation}
The generalized SCMS sequence initiated at $x_0$ is generated by the $m$-fold composition of this operator, denoted $x_m = G_\alpha^m(x_0)$, for $m=1,2,\dots$. 

By imposing appropriate step-size bounds derived from our global constants that are defined in Section \ref{sec:preliminaries}, we establish that the discrete map $G_\alpha$ prevents the sequence from escaping the neighborhood of the ridge. Furthermore, we prove that the magnitude of projected gradient $\xi$ decays at a geometric (linear) rate along the path in the algorithm, which recovers the entire stable ridge, mirroring what is achieved in the continuous setting. 

The following lemma provides the required step-size upper bounds. We define the total arc length of the discrete path generated by the sequence $\{x_m\}_{m=0}^\infty$ as $$\mathcal{L}_\alpha(x_0) = \sum_{m=0}^\infty \|x_{m+1} - x_m\|.$$
\begin{lemma}
\label{lem:geometric_gradient_decay}
Suppose Assumptions (A1) and (A2) hold. Let $\epsilon \in (0, \kappa]$. If the sequence $\{x_m\}_{m=0}^\infty$ is initialized at $x_0 \in \mathcal{R}_{\epsilon}(f)$ and the step size satisfies
\begin{equation}
\label{step size}
    \alpha < \min\left( \frac{\gamma}{L^2}, \; \frac{\gamma}{4 M \epsilon}, \; \frac{d_{\partial}}{\epsilon} \right),
\end{equation}
then for all $m \ge 0$, we have
\begin{enumerate}
    \item[(a)] $x_m \in \mathcal{R}_{\epsilon}(f)$.
    \item[(b)]  $\|\xi(x_{m+1})\| \le \rho \|\xi(x_m)\|$, where $\rho = 1 - \frac{\alpha\gamma}{4} < 1$.
    \item[(c)]  $\mathcal{L}_\alpha(x_0) \le \frac{4}{\gamma} \|\xi(x_0)\|$.
\end{enumerate}
\end{lemma}

With the length of the discrete trajectory bounded and the projected gradient decaying at a geometric rate, we can now show that the generalized SCMS algorithm converges to the stable ridge at a uniform rate.

\begin{theorem}
\label{thm:unified_convergence}
Suppose Assumptions (A1) and (A2) hold, and the step-size condition \eqref{step size} of Lemma \ref{lem:geometric_gradient_decay} is satisfied. For any $\epsilon \in (0, \kappa]$ and any initial point $x_0 \in \mathcal{R}_{\epsilon}(f)$, the sequence $\{x_m\}_{m=0}^\infty$ converges to a limit point $x_\infty \in \mathcal{R}(f)$. Furthermore, the convergence is R-linear uniformly for all starting points $x_0 \in \mathcal{R}_{\epsilon}(f)$, that is, for all $m \ge 0$,
\begin{equation}
\|x_m - x_\infty\| \le C \rho^m,
\end{equation}
where $C = \frac{4\epsilon}{\gamma}$ and $\rho = 1 - \frac{\alpha \gamma}{4}$.
\end{theorem}

Theorem \ref{thm:unified_convergence} guarantees the existence of a well-defined map, which we denote $\Phi_\alpha: \mathcal{R}_{\epsilon}(f) \to \mathcal{R}(f)$, given by the limit of the iterative sequence:
\begin{equation}
\Phi_\alpha(x) = \lim_{m \to \infty} G_\alpha^m(x) , \text{ for } x\in \mathcal{R}_{\kappa}(f).
\end{equation}
To effectively utilize this map in the subsequent finite-sample analysis, we first show that it is continuous.

\begin{proposition}
\label{prop:continuous_limit}
Suppose Assumptions (A1) and (A2) hold, and the step-size condition \eqref{step size} of Lemma \ref{lem:geometric_gradient_decay} is satisfied. For any $\epsilon \in (0, \kappa]$, the limit map $\Phi_\alpha(x)$ is continuous on $\mathcal{R}_{\epsilon}(f)$.
\end{proposition}

Furthermore, it is important to quantify how closely this map $\Phi_\alpha$ based on discrete trajectories approximates $\Phi$ based on continuous flows and defined in \eqref{Phi def}. The next proposition demonstrates that their difference is linearly bounded by the step size $\alpha$.

\begin{proposition}
\label{prop:discrete_limit_error}
Suppose Assumptions (A1) and (A2) hold. Let $\epsilon \in (0, \kappa)$ and $x_0 \in \mathcal{R}_{\epsilon}(f)$. Assume the step size $\alpha$ satisfies the condition \eqref{step size} in Lemma \ref{lem:geometric_gradient_decay} and additionally satisfies
\begin{equation}
    \alpha < \frac{\kappa - \epsilon}{L \epsilon}.
\end{equation}
Then we have
\begin{equation}
    \|\Phi_\alpha(x_0) - \Phi(x_0)\| \le \left( \frac{2 L_\Phi L}{\gamma} \|\xi(x_0)\| \right) \alpha.
\end{equation}
\end{proposition}

Finally, mirroring the surjectivity for the continuous flow established in Theorem \ref{thm:continuous_surjectivity}, we can prove that the generalized SCMS algorithm is also capable of recovering the \textit{entire} stable ridge, when initialized from the outer boundary of the neighborhood. 

\begin{theorem}
\label{thm:discrete_surjectivity}
Suppose Assumptions (A1) and (A2) hold. For $\epsilon \in (0, \kappa]$, let the step size $\alpha > 0$ satisfy
\begin{equation}
    \alpha < \min\left( \frac{\gamma}{L^2}, \; \frac{\gamma}{4 M \epsilon}, \; \frac{1}{L}, \; \frac{d_{\partial}}{\epsilon} \right).
\end{equation}
Then the limit map $\Phi_\alpha$ is surjective from $\partial \mathcal{R}_{\epsilon}(f)$ onto $\mathcal{R}(f)$. That is, for every point $x_* \in \mathcal{R}(f)$, there exists a starting point $x_0 \in \partial \mathcal{R}_{\epsilon}(f)$ such that $\Phi_\alpha(x_0) = x_*$.
\end{theorem}

\section{Statistical Consistency of the Empirical Algorithm}
\label{sec:finite_sample_consistency}

Having characterized the geometric and topological stability of the discrete algorithm based on the population density, we now turn to the practical setting where the true density $f$ is unknown. In practice, given a finite set of independent and identically distributed sample points $X_1, \dots, X_n \sim f$, we can estimate $f$ by using the Kernel Density Estimator (KDE) over $\mathbb{R}^d$:
$$ \widehat{f}(x) = \frac{1}{n h^d} \sum_{i=1}^n K\left(\frac{x - X_i}{h}\right), $$
where $K: \mathbb{R}^d \to \mathbb{R}$ is the smoothing kernel and $h \equiv h_n > 0$ is the bandwidth. 

Using this estimator, we can estimate the gradient $\nabla f(x)$ and Hessian  $\nabla^2 f(x)$ by $\nabla \widehat{f}(x)$ and $\nabla^2 \widehat{f}(x)$, respectively. Let $\widehat{\lambda}_1(x) \ge \dots \ge \widehat{\lambda}_d(x)$ denote the sorted eigenvalues of $\nabla^2 \widehat{f}(x)$. Let $\widehat{E}_\perp(x)$ denote the eigenspace spanned by the eigenvectors corresponding to the smallest $d-k$ eigenvalues of $\nabla^2 \widehat{f}(x)$, and let $\widehat{\Pi}(x)$ denote the orthogonal projection matrix onto this subspace. Then $\xi(x)$ can be estimated by the plug-in estimator
$$ \widehat{\xi}(x) = \widehat{\Pi}(x) \nabla \widehat{f}(x). $$
Furthermore, we define $J_{\widehat{\xi}}(x) = \nabla \widehat{\xi}(x)$ as the Jacobian of $\widehat{\xi}$ at $x$, with its eigenvalues denoted by $\widehat{\mu}_1(x), \dots, \widehat{\mu}_d(x)$, sorted by their real parts in a descending order.

The plug-in stable ridge estimator is
$$ \mathcal{R}(\widehat{f}) = \left\{x \in \mathcal{D} :\; \widehat{\xi}(x) = 0 \text{ and } \mathrm{Re}(\widehat{\mu}_{k+1}(x)) < 0\right\}. $$
For a given threshold $\epsilon > 0$, corresponding to $ \mathcal{R}_{\epsilon}(f)$, we define
$$ \mathcal{R}_{\epsilon}(\widehat{f}) = \left\{x \in \mathcal{D} :\; \|\widehat{\xi}(x)\| \le \epsilon \text{ and } \mathrm{Re}(\widehat{\mu}_{k+1}(x)) < 0 \right\}. $$

Similarly, let $\widehat{\varphi}_t(x)$ denote the trajectory driven by the vector field $\widehat{\xi}$, that is, $$\frac{d}{dt}\widehat{\varphi}_t(x) = \widehat{\xi}(\widehat{\varphi}_t(x)), \quad \text{with } \widehat{\varphi}_0(x) = x.$$ The empirical one-step SCMS operator, parameterized by a constant step size $\alpha > 0$, is defined as
$$ \widehat{G}_\alpha(x) = x + \alpha \widehat{\xi}(x). $$
The sequence for the sample version of the generalized SCMS algorithm is generated by the $m$-fold composition $\widehat{x}_m = \widehat{G}_\alpha^m(\widehat{x}_0)$.

In this section, we derive the statistical consistency and convergence of this empirical algorithm, where the rates of convergence in the error bounds are quantified in terms of the sample size $n$, bandwidth $h$, and step size $\alpha$.

\subsection{Assumptions for Statistical Estimation} 
To guarantee the uniform consistency of $\widehat{f}$ and its higher-order derivatives, we require some regularity conditions on the kernel and bandwidth. For a multi-index $a = (a_1, \dots, a_d)$ of non-negative integers with $|a| = \sum_{i=1}^d a_i = m$, let $D^a f(x) = \frac{\partial^{|a|} f}{\partial x_1^{a_1} \dots \partial x_d^{a_d}}(x)$ denote the corresponding partial derivative of order $m$.

We require the following standard conditions for kernel density estimation and its derivatives. 
\begin{description}
    \item[Assumption (B1).] The kernel $K \in C^4(\mathbb{R}^d, \mathbb{R})$ is a symmetric probability density function with bounded partial derivatives up to the fourth order. It satisfies $\int u u^\top K(u) du = \kappa_2 I$, where $0 < \kappa_2 < \infty$. Furthermore,  
    $$ \mathcal{F}_m = \left\{ x \mapsto D^a K\left(\frac{x - y}{h}\right) : h > 0, y \in \mathbb{R}^d, |a|=m \right\} \quad \text{for } m = 0, \dots, 4, $$
    form Vapnik-Chervonenkis (VC) subgraph classes.
    \item[Assumption (B2).] As the sample size $n \to \infty$, the bandwidth sequence $h_n$ satisfies $h_n \to 0$ and
    $$ \frac{n h_n^{d+8}}{\log n} \to \infty. $$
\end{description}

\begin{remark}
Assumptions (B1) and (B2) are mild for nonparametric estimation of densities and their higher-order derivatives. In (B1), the $C^4$ smoothness guarantees that the vector field $\widehat{\xi}$ is sufficiently smooth on $\mathcal{D}$. The requirement that the families $\mathcal{F}_m$ form VC subgraph classes is standard for the $L_\infty$ stochastic error bounds in empirical process theory \citep{gine2002rates, einmahl2005uniform}. Assumption (B2) places a lower bound on the bandwidth decay rate specifically to ensure that the uniform stochastic error of the fourth-order derivative estimator vanishes uniformly over the domain $\mathcal{D}$. 
\end{remark}

The following lemma provides the uniform convergence rates for the KDE and its derivatives up to the fourth order, which serve as the baseline for all subsequent analyses for the plug-in estimators based on KDE and its derivatives. We restrict our statistical analysis to the compact domain $\mathcal{D}$ given in Assumption (A1).

\begin{lemma}
\label{lem:kde_uniform_convergence_complete}
Under Assumptions (A1), (B1), and (B2), for any constant $c > 0$, there exist positive constants $C_0, C_1, C_2, C_3, C_4$ and a deterministic function $\rho_4(h) = o(1)$ such that for all $n$ sufficiently large, the following bounds hold simultaneously with probability at least $1 - n^{-c}$.
\begin{align}
\max_{|a|=0} \sup_{x \in \mathcal{D}} |D^a \widehat{f}(x) - D^a f(x)| &\le C_0 \left( h^2 + \sqrt{\frac{\log n}{n h^d}} \right)=:C_0 r_0(n), \label{eq:kde_rate_f} \\
\max_{|a|=1} \sup_{x \in \mathcal{D}} |D^a \widehat{f}(x) - D^a f(x)| &\le C_1 \left( h^2 + \sqrt{\frac{\log n}{n h^{d+2}}} \right)=:C_1 r_1(n), \label{eq:kde_rate_g} \\
\max_{|a|=2} \sup_{x \in \mathcal{D}} |D^a \widehat{f}(x) - D^a f(x)| &\le C_2 \left( h^2 + \sqrt{\frac{\log n}{n h^{d+4}}} \right)=:C_2 r_2(n), \label{eq:kde_rate_H} \\
\max_{|a|=3} \sup_{x \in \mathcal{D}} |D^a \widehat{f}(x) - D^a f(x)| &\le C_3 \left( h + \sqrt{\frac{\log n}{n h^{d+6}}} \right)=:C_3 r_3(n), \label{eq:kde_rate_f3} \\
\max_{|a|=4} \sup_{x \in \mathcal{D}} |D^a \widehat{f}(x) - D^a f(x)| &\le C_4 \left(\rho_4(h) + \sqrt{\frac{\log n}{n h^{d+8}}} \right)=:C_4 r_4(n). \label{eq:kde_rate_f4}
\end{align}
\end{lemma}

\subsection{Ridge Regularity for the Estimated Density} 
Recall that the SCMS algorithm operates dynamically based on the density gradient and the eigenspaces of the density Hessian. We need to translate the perturbation bounds for the density derivatives in Lemma \ref{lem:kde_uniform_convergence_complete} into perturbation bounds for the geometric quantities defining SCMS. 

To ensure the sample version of the generalized SCMS algorithm behaves the same as its population version, for any constant $c > 0$, we define $\mathcal{E}_n(c)$ as the event where the uniform convergence bounds in Lemma \ref{lem:kde_uniform_convergence_complete} hold simultaneously. By Lemma \ref{lem:kde_uniform_convergence_complete}, it is guaranteed that $\mathbb{P}(\mathcal{E}_n(c)) \ge 1 - n^{-c}$ for all sufficiently large $n$.

\begin{lemma}
\label{lem:perturbation_bounds}
Under Assumptions (A1)--(A2) and (B1)--(B2), for any $c > 0$, there exists an integer $N_0$ such that for all $n \ge N_0$, conditionally on the event $\mathcal{E}_n(c)$, the following bounds hold.
\begin{enumerate}
    \item[(a)] $\sup_{x \in \mathcal{D}} \max_j |\widehat{\lambda}_j(x) - \lambda_j(x)| \le d C_2 r_2(n)$.
    \item[(b)] $\sup_{x \in \mathcal{D}} \|\widehat{\Pi}(x) - \Pi(x)\| \le 2 d \eta^{-1}C_2 r_2(n).$
    \item[(c)] There exists a constant $C_\xi > 0$ such that 
    $$ \sup_{x \in \mathcal{D}} \|\widehat{\xi}(x) - \xi(x)\| \le \epsilon_\xi(n) := C_\xi r_2(n). $$
    \item[(d)] There exists a constant $C_J > 0$ such that 
    $$ \sup_{x \in \mathcal{D}} \|J_{\widehat{\xi}}(x) - J_\xi(x)\| \le \epsilon_J(n) := C_J r_3(n). $$
\end{enumerate}
\end{lemma}

Consequently, for a sufficiently large sample size, the geometric quantities defining SCMS based on $\widehat{f}$ are stochastically close to those based on the true density $f$. The following corollary formalizes that $\widehat{f}$ satisfies the ridge-regular condition for $f$ as defined in Definition \ref{def:function_class}, with parameters scaled by a factor of $1/2$ to absorb the uniform stochastic errors.

\begin{corollary}
\label{cor:empirical_assumptions}
Assume Assumptions (A1)--(A2) and (B1)--(B2) hold with $f \in \mathcal{F}(\eta, \gamma, \kappa, \mathcal{D})$. For any $c > 0$, there exists an integer $N_1 \ge N_0$ such that for all $n \ge N_1$, conditionally on $\mathcal{E}_n(c)$, $\widehat{f}$ satisfies Assumption (A1) and
$$ \widehat{f} \in \mathcal{F}\left(\frac{\eta}{2}, \frac{\gamma}{2}, \frac{\kappa}{2}, \mathcal{D}\right). $$
Specifically, this guarantees the following.
\begin{enumerate}
    \item[(a)] $\widehat{f} \in C^4(\mathbb{R}^d, \mathbb{R})$, and there exist constants $B_\ell > 0$ such that $\max_{|a|=\ell} \sup_{x \in \mathcal{D}} |D^a \widehat{f}(x)| \le B_\ell$, for $\ell=0,\dots,4$.
    \item[(b)] $\widehat{\lambda}_k(x) - \widehat{\lambda}_{k+1}(x) \ge \frac{\eta}{2} > 0$ for all $x \in \mathcal{D}$.
    \item[(c)] For all $x \in \mathcal{R}_{\kappa/2}(\widehat{f})$, 
    $$ \sup\{ u^\top J_{\widehat{\xi}}(x) u: u \in \widehat{E}_{\perp}(x), \; \|u\| = 1 \} \le -\frac{\gamma}{2} < 0. $$
    \item[(d)] $\inf \big\{ \|\widehat{\xi}(x)\| :\; x \in \mathcal{D}, \; \mathrm{Re}(\widehat{\mu}_{k+1}(x)) \ge -\frac{\gamma}{4} \big\} \ge \frac{\kappa}{2} > 0.$
\end{enumerate}
\end{corollary}

\subsection{Consistency of the Plug-in Ridge Estimator} 
Before analyzing discrete trajectories induced by $\widehat{G}_\alpha$, which are expected to converge to the plug-in ridge estimator $\mathcal{R}(\widehat{f})$, we first confirm that the target $\mathcal{R}(\widehat{f})$ is close to the true ridge $\mathcal{R}(f)$. 

The following theorem demonstrates that the Hausdorff distance between the estimated ridge and the true ridge shares the same rate of convergence as the estimation of the Hessian.

\begin{theorem}
\label{thm:hausdorff_consistency}
Assume Assumptions (A1)--(A2) and (B1)--(B2) hold. For any $c > 0$, there exists an integer $N_2 \ge N_1$ such that for all $n \ge N_2$, conditionally on the event $\mathcal{E}_n(c)$,
\begin{equation}
    d_H(\mathcal{R}(\widehat{f}), \mathcal{R}(f)) \le \frac{2\epsilon_\xi(n)}{\gamma} = O\left( h^2 + \sqrt{\frac{\log n}{n h^{d+4}}} \right).
\end{equation}
\end{theorem}

\begin{remark}
\label{rem:optimal_rate}
Minimizing the Hausdorff error rate in Theorem \ref{thm:hausdorff_consistency} requires balancing the squared bias rate $O(h^2)$ with the uniform stochastic error rate $O\left(\sqrt{\log n / (n h^{d+4})}\right)$. This produces the balancing bandwidth $h \asymp (\log n / n)^{\frac{1}{d+8}}$ such that $\frac{n h^{d+8}}{\log n} \asymp 1$, which however narrowly fails the requirement in Assumption (B2) for the uniform convergence of the fourth derivatives. To address this, we apply an arbitrarily small logarithmic correction. For any constant $\delta > 0$, choosing the bandwidth
$$ h \asymp \left( \frac{(\log n)^{1+\delta}}{n} \right)^{\frac{1}{d+8}} $$
ensures that $\frac{n h^{d+8}}{\log n} \asymp (\log n)^\delta \to \infty$, which also satisfies Assumption (B2). Under this choice, the estimator is slightly oversmoothed, and the Hausdorff error is dominated by the bias, which yields
$$ d_H(\mathcal{R}(\widehat{f}), \mathcal{R}(f)) = O\left( \left( \frac{(\log n)^{1+\delta}}{n} \right)^{\frac{2}{d+8}} \right). $$
\end{remark}

\subsection{Convergence of the Generalized SCMS Algorithm: Sample Version} 
We now focus on the extraction of the estimated ridge $\mathcal{R}(\widehat{f})$ using the empirical SCMS operator $\widehat{G}_\alpha(x) = x + \alpha \widehat{\xi}(x)$. For any $\epsilon \in (0, \kappa/2]$, we define the boundary of $\mathcal{R}_{\epsilon}(\widehat{f})$ as
\begin{equation}
\partial \mathcal{R}_{\epsilon}(\widehat{f}) = \big\{x \in \mathcal{D} :\; \|\widehat{\xi}(x)\| = \epsilon \text{ and } \mathrm{Re}(\widehat{\mu}_{k+1}(x)) < 0 \big\}.
\end{equation}
Let $\widehat{\mathcal{X}}_m = \{ \widehat{G}_\alpha^m(x) : x \in \partial \mathcal{R}_{\epsilon}(\widehat{f}) \}$ denote the image of this boundary after $m$ iterations of $\widehat{G}_\alpha$.

The following theorem mirrors the population result in Theorem \ref{thm:unified_convergence}, proving that the sample version of the generalized SCMS algorithm uniformly converges to the estimated ridge at an R-linear rate.

\begin{theorem}
\label{thm:empirical_scms_convergence}
Let $\epsilon \in (0, \kappa/2]$. Under Assumptions (A1)--(A2) and (B1)--(B2), for any $c > 0$, there exists an integer $N_2 \ge N_1$ such that for all $n \ge N_2$, conditionally on the event $\mathcal{E}_n(c)$, the following holds.

For any point $\widehat{x}_0 \in \partial \mathcal{R}_{\epsilon}(\widehat{f})$, if the step size $\alpha > 0$ satisfies
\begin{equation}
    \alpha < \min\left( \frac{\gamma}{8 L^2}, \; \frac{\gamma}{16 M \epsilon}, \; \frac{d_{\partial}}{\epsilon} \right),
\end{equation}
then the sequence $\{\widehat{x}_m = \widehat{G}_\alpha^m(\widehat{x}_0)\}_{m=0}^\infty$ converges to a limit point $\widehat{x}_\infty \in \mathcal{R}(\widehat{f})$, and
\begin{equation}
    \|\widehat{x}_m - \widehat{x}_\infty\| \le \frac{8 \epsilon}{\gamma} \left(1 - \frac{\alpha \gamma}{8}\right)^m, \text{ for all } m\ge0.
\end{equation}
\end{theorem}

Just as in the population setting, it is insufficient to characterize the behavior of the sample version of the generalized SCMS algorithm by pointwise convergence alone, as it is possible that the algorithm only recovers an arbitrary fraction of the stable ridge. We need to demonstrate topological surjectivity onto the estimated stable ridge to ensure a full recovery of the target.

\begin{theorem}
\label{thm:empirical_discrete_surjectivity}
Let $\epsilon \in (0, \kappa/2]$. Under Assumptions (A1)--(A2) and (B1)--(B2), for any $c > 0$, there exists an integer $N_2 \ge N_1$ such that for all $n \ge N_2$, conditionally on the event $\mathcal{E}_n(c)$, the following holds. 

If the step size $\alpha > 0$ satisfies
\begin{equation}
    \alpha < \min\left( \frac{\gamma}{8 L^2}, \; \frac{\gamma}{16 M \epsilon}, \; \frac{1}{2L}, \; \frac{d_{\partial}}{\epsilon} \right),
\end{equation}
then the limit map $\widehat{\Phi}_\alpha(x) = \lim_{m \to \infty} \widehat{G}_\alpha^m(x)$ is surjective from the boundary set $\partial \mathcal{R}_{\epsilon}(\widehat{f})$ onto the estimated ridge $\mathcal{R}(\widehat{f})$.
\end{theorem}

\begin{corollary}
\label{cor:algorithmic_hausdorff}
Assume the conditions of Theorem \ref{thm:empirical_discrete_surjectivity} hold. For any $c > 0$, there exists an integer $N_2$ such that for all $n \ge N_2$, with probability at least $1 - n^{-c}$, we have
\begin{equation}
    d_H(\widehat{\mathcal{X}}_m, \mathcal{R}(\widehat{f})) \le \frac{8\epsilon}{\gamma} \left(1 - \frac{\alpha \gamma}{8}\right)^m.
\end{equation}
\end{corollary}

We are now ready to state the capstone result of this section. By combining Theorem \ref{thm:hausdorff_consistency} and Corollary \ref{cor:algorithmic_hausdorff} and using a triangle inequality for the Hausdorff distance, we provide the total error bound for ridge estimation using the sample version of the generalized SCMS algorithm.

\begin{theorem}
\label{thm:total_error_bound}
Assume the conditions of Theorem \ref{thm:empirical_discrete_surjectivity} hold. For any $c > 0$, there exists an integer $N_2$ such that for all $n \ge N_2$, with probability at least $1 - n^{-c}$, we have
\begin{equation}
\label{total_error_bound}
    d_H(\widehat{\mathcal{X}}_m, \mathcal{R}(f)) \le \frac{8\epsilon}{\gamma} \left(1 - \frac{\alpha \gamma}{8}\right)^m + \frac{2 \epsilon_\xi(n)}{\gamma}.
\end{equation}
\end{theorem}

\begin{remark}
\label{rem:optimal_stopping_time}
In Theorem \ref{thm:total_error_bound}, the total error in estimating the population stable ridge is decomposed into the computational error and the statistical error, corresponding to the first and second terms on the right side of \eqref{total_error_bound}, respectively. The computational error vanishes exponentially fast with the number of iterations $m$, while the statistical error depends on the sample size $n$ and bandwidth $h$. 

To achieve the convergence rate for ridge estimation carefully balanced in Remark \ref{rem:optimal_rate}, we can execute the operator $\widehat{G}_\alpha$ until the computational error is dominated by the statistical error. Using the bandwidth $h \asymp \big(\frac{(\log n)^{1+\delta}}{n}\big)^{\frac{1}{d+8}}$, the statistical error is bounded by $\epsilon_\xi(n) = O\Big( \big(\frac{(\log n)^{1+\delta}}{n}\big)^{\frac{2}{d+8}} \Big)$, following from Lemma \ref{lem:perturbation_bounds}. 

By setting the computational error $\frac{8\epsilon}{\gamma} (1 - \alpha \gamma / 8)^m \asymp \epsilon_\xi(n)$, we find the required number of iterations before stopping the algorithm is
\begin{equation}\label{eq:stopping_time_approx}
m^* =  \left\lceil \frac{\log(8\epsilon/\gamma) - \log(\epsilon_\xi(n))}{-\log(1 - \alpha \gamma / 8)} \right\rceil = O\left( \log \left( \frac{1}{\epsilon_\xi(n)} \right) \right) = O(\log n).
\end{equation}

Evaluating at this stopping time, we obtain the rate of convergence for the total error using the sample version of the generalized SCMS algorithm as
$$ d_H(\widehat{\mathcal{X}}_{m^*}, \mathcal{R}(f)) = O\left( \left( \frac{(\log n)^{1+\delta}}{n} \right)^{\frac{2}{d+8}} \right). $$
\end{remark}

\section{Convergence and Computational Complexity of the Original SCMS Algorithm}
\label{sec:original_scms_log_density}

The generalized SCMS framework developed in Sections \ref{sec:discrete_optimization} and \ref{sec:finite_sample_consistency} operates on the density $f$ and its estimator $\widehat{f}$ using a constant step size $\alpha$. We now extend the theory there to the original SCMS algorithm \citep{ozertem2011locally}, which is based on an estimator of the log-density and couples its step size to the smoothing bandwidth.

Define the log-density as $p(x) = \log f(x)$ and its plug-in estimator as $\widehat{p}(x) = \log \widehat{f}(x)$. While the gradients of the density and log-density have the same direction ($\nabla p(x) = \frac{1}{f(x)}\nabla f(x)$), their Hessians have different eigenspaces in general: 
\begin{equation}\label{eq:log_density_hessian}
\nabla^2 p(x) = \frac{\nabla^2 f(x)}{f(x)} - \frac{\nabla f(x)\nabla f(x)^\top}{f(x)^2}.
\end{equation}
This makes the log-density ridge $\mathcal{R}(p)$ sometimes distinct from the density ridge $\mathcal{R}(f)$. As discussed in Section \ref{sec:introduction}, the theoretical target of the original SCMS algorithm is the log-density ridge $\mathcal{R}(p)$.

To establish the convergence of the original SCMS algorithm, we update Assumption (A2) to
\begin{description}
    \item[Assumption (A2$^\prime$).] 
$p \in \mathcal{F}(\eta_{\log}, \gamma_{\log}, \kappa_{\log}, \mathcal{D})$ for some positive constants $\eta_{\log}, \gamma_{\log}$, and $\kappa_{\log}$.
\end{description}
Furthermore, in (B1), we additionally assume that the smoothing kernel $K$ is radially symmetric. Specifically, we assume there exists a differentiable \textit{profile function} $\textsc{k}: [0, \infty) \to [0, \infty)$ such that the kernel can be written as
$$K(x) = c_{\textsc{k}} \textsc{k}(\|x\|^2),$$
where $c_{\textsc{k}} > 0$ is a normalization constant such that $K$ integrates to 1 over $\mathbb{R}^d$. We require that $\textsc{k}(u)$ is non-increasing for all $u \ge 0$, and strictly decreasing for all $u$ such that $\textsc{k}(u) > 0$. The updated Assumption (B1) is called Assumption (B1$^\prime$). We retain Assumptions (A1) and (B2). Because the profile function is non-increasing, its negative derivative yields a non-negative weight profile $\textsc{h}(u) = -\textsc{k}'(u) \ge 0$, which is positive on the support of the kernel (i.e., wherever $\textsc{k}(u) > 0$). We similarly define $c_{\textsc{h}} > 0$ as the corresponding normalization constant for $\textsc{h}$.

Let $\Pi^{\log}(x)$ denote the orthogonal projection matrix onto the eigenspace associated with the smallest $d-k$ eigenvalues of the log-density Hessian $\nabla^2 p(x)$. Let $\widehat{\Pi}^{\log}(x)$ denote its plug-in estimator when $\nabla^2 p(x)$ is replaced by $\nabla^2 \widehat{p}(x)$. Denote $\xi^{\log}(x) = \Pi^{\log}(x)\nabla p(x)$, and its Jacobian as $J_{\xi^{\log}}(x) = \nabla \xi^{\log}(x)$. Similarly, denote $\widehat{\xi}^{\log}(x) = \widehat{\Pi}^{\log}(x) \nabla \widehat{p}(x)$ and let $J_{\widehat{\xi}^{\log}}(x)$ be its Jacobian. 

To formally introduce the original SCMS algorithm, we first define the Mean Shift vector 
\begin{equation}\label{eq:mean_shift_vector}
{\sf MS}_{n,h}(x) = \frac{\sum_{i=1}^n X_i \, \textsc{h}\big(\big\|\frac{x - X_i}{h}\big\|^2\big)}{\sum_{i=1}^n \textsc{h}\big(\big\|\frac{x - X_i}{h}\big\|^2\big)} - x,
\end{equation}
which computes a locally weighted shift pointing toward the center of mass of the neighboring data points around $x$, using the profile $\textsc{h}$. The original SCMS algorithm generates a sequence of iterates $\{x_m\}_{m=0}^\infty$ via the update rule determined by $\widehat{\Pi}^{\log}{\sf MS}_{n,h}$, as specified in \eqref{eq:scms_discrete_update}. 

In what follows, we show that the original SCMS algorithm can be represented in the form of 
\begin{equation}
\label{original SCMS form}
x_{m+1} = x_m +  \alpha_n(x_m) \widehat{\xi}^{\log}(x_m),
\end{equation}
where $\alpha_n$ is a varying step size depending on the bandwidth $h$. 
 
 By defining $\widehat{q}(x) = \frac{c_{\textsc{h}}}{n h^d} \sum_{i=1}^n \textsc{h}\big(\big\|\frac{x - X_i}{h}\big\|^2\big)$ as another kernel density estimator using the profile $\textsc{h}$, we can write
\begin{equation}\label{eq:mean_shift_log_gradient}
{\sf MS}_{n,h}(x) = C_K h^2 \frac{\nabla \widehat{f}(x)}{\widehat{q}(x)} = \alpha_n(x) \nabla \widehat{p}(x),
\end{equation}
where $C_K = \frac{c_{\textsc{h}}}{2 c_{\textsc{k}}}$ is a constant only depending on the kernel, and $\alpha_n(x) = C_K h^2 \frac{\widehat{f}(x)}{\widehat{q}(x)}$. This reveals that the Mean Shift vector is parallel to the log-density gradient estimator $\nabla \widehat{p}(x)$, scaled by an adaptive step size $\alpha_n(x)$. Notably, when utilizing the standard Gaussian kernel, which is the primary focus of the original SCMS \citep{ozertem2011locally}, the profiles satisfy $\textsc{k}(u) = e^{-u/2}$ and $\textsc{h}(u) = \frac{1}{2}e^{-u/2}$. At the same time, $c_{\textsc{h}} = 2 c_{\textsc{k}}$ (yielding $C_K = 1$), and the density estimators are identical ($\widehat{q}(x) \equiv \widehat{f}(x)$). In this special case, the step size $\alpha_n(x)$ becomes independent of $x$, that is, $\alpha_n(x) \equiv h^2$.

Plugging \eqref{eq:mean_shift_log_gradient} into \eqref{eq:scms_discrete_update}, we immediately recover the form of the original SCMS in \eqref{original SCMS form}, or equivalently, $x_{m+1} = \widehat{G}_{\alpha_n}^{\log}(x_m)$, where
\begin{equation}\label{eq:scms_log_density_update}
\widehat{G}_{\alpha_n}^{\log}(x) = x + \alpha_n(x) \widehat{\xi}^{\log}(x).
\end{equation}
This one-step operator for the original SCMS is analogous to $\widehat{G}_\alpha$ for the sample version of the generalized SCMS, and analyses are also similar to those in Section \ref{sec:finite_sample_consistency}. 

First, similar to Lemma \ref{lem:kde_uniform_convergence_complete}, under Assumptions (A1), (B1$^\prime$) and (B2), for any constant $c > 0$, we have $\mathbb{P}(\mathcal{E}_n^\prime(c)) \ge 1 - n^{-c}$ for $n$ large enough, where $\mathcal{E}_n^\prime(c)$ is the event such that 
\begin{equation*}
\max_{|a|=\ell} \sup_{x \in \mathcal{D}} |D^a \widehat{p}(x) - D^a p(x)| \le C_\ell^{\log} r_\ell(n) \quad \text{for } \ell=0, 1, \dots, 4,
\end{equation*}
where $r_\ell(n)$ are the rates from Lemma \ref{lem:kde_uniform_convergence_complete}, and $C_\ell^{\log}$ are positive constants. This can be verified by noticing $f(x) \ge c_0 > 0$ under Assumption (A1), and using a telescoping argument. Consequently, results similar to Lemma \ref{lem:perturbation_bounds} and Corollary \ref{cor:empirical_assumptions} in the log-density setting hold. In particular, for the perturbation of the vector field, conditionally on $\mathcal{E}_n(c)$ we have
\begin{equation*}
\sup_{x \in \mathcal{D}} \|\widehat{\xi}^{\log}(x) - \xi^{\log}(x)\| \le \epsilon_\xi^{\log}(n),
\end{equation*}
where $\epsilon_\xi^{\log}(n) = O(r_2(n)) = O\Big( h^2 + \sqrt{\frac{\log n}{n h^{d+4}}} \Big)$. 

Furthermore, since both $\widehat{f}$ and $\widehat{q}$ are valid kernel density estimators, they converge uniformly to the population density $f$, and hence their ratio converges uniformly to $1$ across the domain $\mathcal{D}$, where $f(x) \ge c_0 > 0$. We define the ratio error as
\begin{equation}\label{eq:scms_ratio_error}
\Delta_n = \sup_{x \in \mathcal{D}} \Big| \frac{\widehat{f}(x)}{\widehat{q}(x)} - 1 \Big|.
\end{equation}
Conditionally on $\mathcal{E}_n(c)$, $\Delta_n \to 0$ as $n \to \infty$, and hence $\alpha_n(x) \asymp h^2$ uniformly in $\mathcal{D}$, ensuring the adaptive step size remains stable.

For $\epsilon \in (0, \kappa_{\log}/2]$, we define the boundary set of $\mathcal{R}_{\epsilon}(\widehat{p})$ as
\begin{equation*}
    \partial \mathcal{R}_{\epsilon}(\widehat{p}) = \big\{x \in \mathcal{D} :\; \|\widehat{\xi}^{\log}(x)\| = \epsilon \text{ and } \mathrm{Re}(\widehat{\mu}_{k+1}^{\log}(x)) < 0 \big\}.
\end{equation*}
The following lemma guarantees that for a sufficiently large sample size $n$ (which also ensures a sufficiently small adaptive step size $\alpha_n$), the original SCMS algorithm can fully recover the estimated log-density stable ridge $\mathcal{R}(\widehat{p})$ if starting from the boundary $\partial \mathcal{R}_{\epsilon}(\widehat{p})$ with high probability.

\begin{lemma}
\label{lem:automatic_stability}
Let $\epsilon \in (0,\kappa_{\log}/2]$. Under Assumptions (A1), (A2$^\prime$), (B1$^\prime$), and (B2), for any $c > 0$, there exists an integer $N_3$ such that for all $n \ge N_3$, conditionally on the event $\mathcal{E}_n(c)$, the map $\widehat{G}_{\alpha_n}^{\log}$ defined in \eqref{eq:scms_log_density_update} is a local diffeomorphism on $\mathcal{R}_\epsilon(\widehat{p})$, and its limit map $\widehat{\Phi}^{\log} = \lim_{m \to \infty} (\widehat{G}^{\log})^m$ exists and is surjective from the boundary set $\partial \mathcal{R}_{\epsilon}(\widehat{p})$ onto the empirical log-density ridge $\mathcal{R}(\widehat{p})$.
\end{lemma}

Let $\widehat{\mathcal{X}}_m^{\log} = \big\{ (\widehat{G}^{\log})^m(x) : x \in \partial \mathcal{R}_{\epsilon}(\widehat{p}) \big\}$ denote the image of the boundary set after running the original SCMS algorithm for $m$ steps. The final theorem below provides the total convergence bounds for the original SCMS algorithm.

\begin{theorem}
\label{thm:scms_total_convergence}
Let $\epsilon \in (0,\kappa_{\log}/2]$. Under Assumptions (A1), (A2$^\prime$), (B1$^\prime$), and (B2), for any $c > 0$, there exists an integer $N_3$ such that for all $n \ge N_3$, with probability at least $1 - n^{-c}$, we have
    \begin{equation}
    \label{log ridge rate}
        d_H(\mathcal{R}(\widehat{p}), \mathcal{R}(p)) \le \frac{2 \epsilon_\xi^{\log}(n)}{\gamma_{\log}},
    \end{equation}
 and 
 \begin{equation}
 \label{original SCMS speed}
    d_H(\widehat{\mathcal{X}}_m^{\log}, \mathcal{R}(\widehat{p})) \le \frac{8\epsilon}{\gamma_{\log}} \rho_n^m,
\end{equation}
where $\rho_n = 1 - \frac{C_K \gamma_{\log}}{16} h^2 $, and consequently,
     \begin{equation}
        d_H(\widehat{\mathcal{X}}_m^{\log}, \mathcal{R}(p)) \le \frac{8\epsilon}{\gamma_{\log}} \rho_n^m + \frac{2 \epsilon_\xi^{\log}(n)}{\gamma_{\log}}.
    \end{equation}
\end{theorem}

\begin{remark}
\label{rem:original_stopping_time}
The result in Theorem \ref{thm:scms_total_convergence} exposes the severe computational cost of the original SCMS algorithm, for which the step size $\alpha_n$ is proportional to $h^2$, as the sample size $n \to \infty$ and the bandwidth $h \to 0$. 

Typically, the stopping criterion for the original SCMS algorithm can be set as $\|\widehat{\xi}^{\log}(x)\| < \epsilon_{\min}$ for a small value of $\epsilon_{\min}>0$. This implies $\widehat{\mathcal{X}}_m^{\log}\subset \mathcal{R}_{\epsilon_{\min}}(\widehat{p})$ and hence $d_H(\widehat{\mathcal{X}}_m^{\log}, \mathcal{R}(\widehat{p})) =O( \epsilon_{\min})$ using an empirical result similar to Corollary \ref{cor:hausdorff_bound}. By invoking \eqref{original SCMS speed} and applying the inequality $\log(1 - x) \le -x$ for $x \in (0,1)$, we find the required number of iterations is $O(h^{-2})$. 

To achieve the targeted convergence rate in \eqref{log ridge rate} for nonparametric estimation, we need to iterate the original SCMS algorithm until the computational error $\frac{8\epsilon}{\gamma_{\log}} \rho_n^m$ is dominated by the statistical error $\epsilon_\xi^{\log}(n)= O\Big( h^2 + \sqrt{\frac{\log n}{n h^{d+4}}} \Big)$. Setting the two errors equal yields a required number of iterations:
\begin{equation}\label{eq:stopping_time_approx_log}
m^* = \left\lceil \frac{\log(8\epsilon/\gamma_{\log}) - \log(\epsilon_\xi^{\log}(n))}{C_K h^2 \gamma_{\log} / 16} \right\rceil = O\left( \frac{1}{h^2} \log \left( \frac{1}{\epsilon_\xi^{\log}(n)} \right) \right).
\end{equation}
Because the log-density estimate shares the same uniform convergence rates as the KDE, we substitute the bandwidth $h \asymp \Big( \frac{(\log n)^{1+\delta}}{n} \Big)^{\frac{1}{d+8}}$ derived in Remark \ref{rem:optimal_rate} to satisfy Assumption (B2). Then the required number of iterations blows up polynomially with the sample size:
\begin{equation}
m^* = O\left( n^{\frac{2}{d+8}} (\log n)^{1 - \frac{2(1+\delta)}{d+8}} \right).
\end{equation}

The high computational complexity is mainly due to the Mean Shift operations implicitly using step size $\alpha_n \asymp h^2$. Our generalized SCMS framework developed in Section \ref{sec:discrete_optimization} and Section \ref{sec:finite_sample_consistency} offers a solution by applying a constant step size $\alpha > 0$:
\begin{equation}
\label{generalizedSCMSlog}
x_{m+1} = x_m + \alpha \widehat{\xi}^{\log}(x_m).
\end{equation}
Similar to Remark \ref{rem:optimal_stopping_time}, when the step size $\alpha$ is appropriately calibrated, with the log-density ridge $\mathcal{R}(p)$ as the estimation target, the generalized SCMS algorithm in \eqref{generalizedSCMSlog} achieves the statistical error rate only requiring $O(\log n)$ iterations. This reduced computational complexity makes log-density ridge extraction feasible when the sample size is massive.
\end{remark}

\section{Numerical Experiments}
\label{sec:numerical_experiments}

To validate the theoretical results regarding the computational complexity of SCMS algorithms and their statistical consistency in estimating the log-density stable ridge, we conduct Monte Carlo simulations based on a probability density $f$, which is constructed by convolving the uniform distribution supported on $\mathcal{M}$ and an isotropic Gaussian distribution $\mathcal{N}(0, \sigma^2 I_2)$ with $\sigma = 0.5$, where $\mathcal{M}$ is the 1-dimensional circle of radius $r = 2$ centered at the origin in $\mathbb{R}^2$. The density $f$ is radially symmetric and can be expressed as $f(x) = g(\|x\|)$ for a function $g: [0, \infty) \to [0, \infty)$. Such a model is further discussed in Appendix \ref{Radial Symmetry}. 

Recall that $p$ denotes the log density $\log f$. For this density, the stable log-density ridge is $\mathcal{R}_{\text{stable}}(p) = \{x\in\mathbb{R}^2: \|x\| =r^*\}$, where $r^* \approx 1.93$. For theoretical purposes, we can find a tubular neighborhood $\mathcal{D} = \{x \in \mathbb{R}^2 : \rho_{\min} \le \|x\|_2 \le \rho_{\max}\}$ that satisfies Assumption (A2$^\prime$). This requires $(\log g)'' < 0$ on this region, or roughly $0.42 \lessapprox \rho_{\min} < r^* < \rho_{\max} <\infty$. 

From this model, we draw i.i.d. random samples $X_i = (r \cos(\Theta_i) + U_{i}, r \sin(\Theta_i) + V_{i}) \in \mathbb{R}^2$, for $i=1,\dots,n$, where $\Theta_i \sim \text{Uniform}(0, 2\pi)$, and $(U_i,V_i)\sim \mathcal{N}(0, 0.5^2 I_2)$ with $\Theta_i$ independent of $(U_i,V_i)$. The sample sizes $n$ range in $\{10^3, 10^{3.5}, 10^4, 10^{4.5}, 10^5\}$. To satisfy Assumption (B2) (also see Remark \ref{rem:optimal_stopping_time}), we choose the bandwidth $h = C (n^{-1} \log^{1+\delta} n)^{1/(d+8)}$, with $d=2$ and $\delta = 0.1$, where the constant $C$ is calibrated such that $h = 0.3$ for $n = 10^3$. We use the Gaussian kernel throughout the experiments. Our goal is to estimate $\mathcal{R}(p)$ by extracting $\mathcal{R}(\widehat{p})$.

To examine the trajectories of the original and generalized SCMS algorithms, the initial points $x_0$ are selected at $(u, 0)\in\mathbb{R}^2$ for $u \in \{0.1, 1.5, 3.0\}$. Specifically, $x_0\in \mathcal{D}$ when $u=1.5$ and $u=3.0$, while the initialization at $u=0.1$ forces the algorithms to navigate for some distance on the density landscape before entering $\mathcal{D}$. The stopping criterion for all the SCMS algorithms is set to be $\|\widehat{\xi}^{\log}(x)\| < 10^{-6}$.

We calculate the required number of iterations for the SCMS algorithms and their statistical errors by averaging over 15 independent replications. As evidenced by the narrow standard error ribbons in Figure \ref{fig:convergence} and Figure \ref{fig:consistency} below, this number of replications is sufficient to make stable asymptotic conclusions.

\subsection{Computational Complexity and Algorithmic Stability}

\begin{figure}[htbp]
    \centering
    \includegraphics[width=\textwidth]{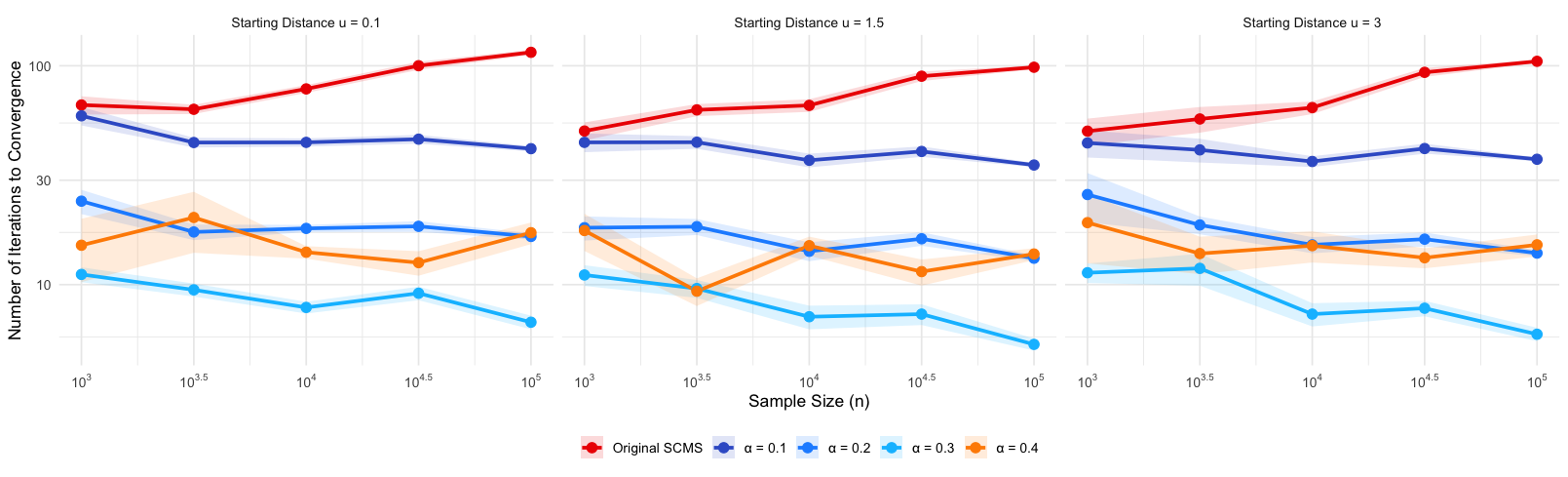}
    \caption{Number of iterations to convergence across varying sample sizes $n$ and starting distances $u$ for the original SCMS and Generalized SCMS with step sizes $\alpha=0.1,0.2,0.3,0.4$. The plot axes are visualized on a log-log scale.}
    \label{fig:convergence}
\end{figure}

First, we want to verify the computational complexity of the SCMS algorithms discussed in Remark \ref{rem:original_stopping_time}. As illustrated in Figure \ref{fig:convergence}, the iteration count for the original SCMS exhibits an approximate linear trend on the log-log scale. Remark \ref{rem:original_stopping_time} predicts an approximate iteration complexity scaling of $O(h^{-2})$, which is $O(n^{2/(d+8)}) = O(n^{0.2})$ up to a logarithm factor given $d = 2$. To only evaluate large sample asymptotic behavior, we fit a linear regression based on the logarithms of the mean number of iterations and the logarithms of the sample sizes $n$ for $n\in\{10^4, 10^{4.5}, 10^5\}$. For $u \in \{1.5, 3.0\}$ (corresponding to $x_0\in\mathcal{D}$), the fitted log-log slopes are $0.174$ and $0.211$, respectively, well approximating the theoretical bound. Even for $u=0.1$ (corresponding to $x_0\notin\mathcal{D}$), the fitted log-log slope is $0.167$. 

In contrast, evaluating the generalized SCMS across a range of step sizes $\alpha \in \{0.1, 0.2, 0.3, 0.4\}$ reveals a computational paradigm shift. For example, focusing on $\alpha = 0.2$, the fitted log-log slopes for $u \in \{0.1, 1.5, 3.0\}$ are $-0.037$, $-0.032$, and $-0.038$, respectively, which are approximately zero. The slight negative trend even underscores the higher efficiency of the algorithm as the sample size grows. This confirms that decoupling the step size from the bandwidth in the generalized SCMS algorithm bypasses the $O(h^{-2})$ computational bottleneck for the original SCMS algorithm, as discussed in Remark \ref{rem:original_stopping_time}. We highlight that similar behaviors are observed between $u \in \{1.5, 3.0\}$ and $u=0.1$, just like for the original SCMS algorithm. This indicates that our theoretical analyses requiring assumptions on $\mathcal{D}$ can be extended to regions beyond $\mathcal{D}$. See the Discussion section.

\begin{figure}[htbp]
    \centering
    \includegraphics[width=0.9\textwidth]{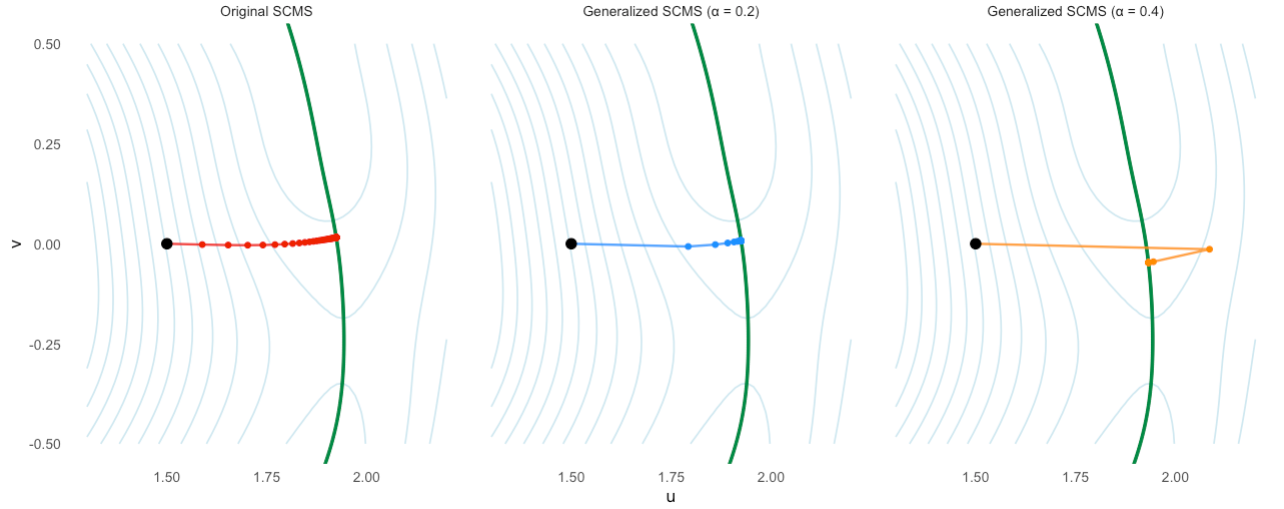}
    \caption{Trajectories of the original and generalized SCMS algorithms. The black dot denotes the starting point and the solid green line is the estimated stable ridge $\mathcal{R}(\widehat{p})$. Setting $\alpha=0.4$ for the generalized SCMS results in overshooting and underdamped oscillation across the estimated ridge.}
    \label{fig:overshoot}
\end{figure}

While the constant step size for the generalized SCMS algorithm has computational benefits compared to the original SCMS algorithm, $\alpha$ needs to be carefully calibrated. As shown in Figure \ref{fig:convergence}, while increasing the step sizes $\alpha$ from 0.1 to 0.3 can speed up the computation for the generalized SCMS, increasing the step size $\alpha$ from 0.3 to 0.4 instead requires more iterations for convergence. An even larger step size could lead to divergence. This is similar to the well-known overshooting problem in gradient ascent algorithms. Therefore, the step size $\alpha$ needs to be carefully tuned, usually by the trial and error method. On the other hand, there is no such a step-size selection problem for the original SCMS algorithm. Figure \ref{fig:overshoot} visualizes the trajectories of the original SCMS algorithm and the generalized SCMS algorithm with $\alpha=0.2$ and $0.4$, where it can be seen that the original SCMS requires an excessive sequence of small steps before reaching the estimated ridge, while the generalized SCMS with $\alpha=0.4$ exhibits a zig-zagging trajectory which overshoots the target during the early stage of the algorithm. 

\subsection{Verification of Statistical Consistency}

\begin{figure}[htbp]
    \centering
    \includegraphics[width=\textwidth]{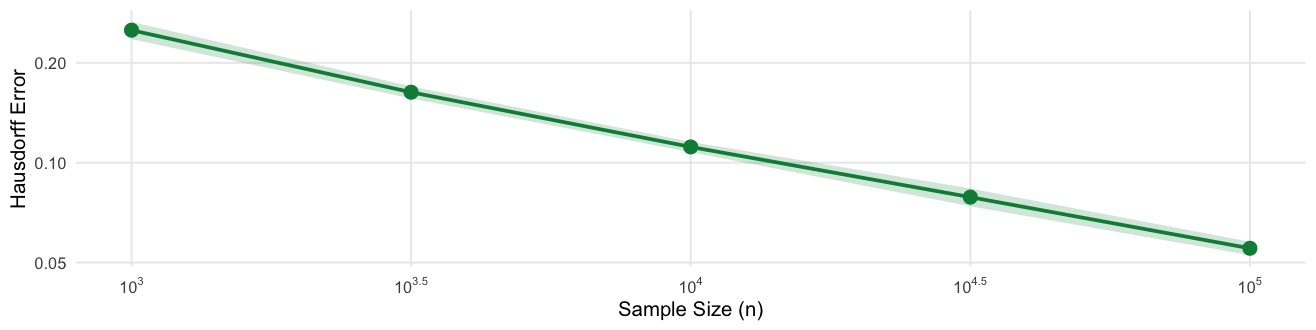}
    \caption{Hausdorff distance between the estimated ridge $\mathcal{R}(\widehat{p})$ and the population ridge $\mathcal{R}(p)$ across varying sample sizes $n$. The plot axes are visualized on a log-log scale.}
    \label{fig:consistency}
\end{figure}

Next, we validate the statistical consistency of the ridge's plug-in estimator by evaluating the Hausdorff distance between the estimated ridge $\mathcal{R}(\widehat{p})$ and the population ridge $\mathcal{R}(p)$. Here, $\mathcal{R}(\widehat{p})$ is extracted using the generalized SCMS algorithm with step size $\alpha = 0.2$. Because we used the stopping criterion $\|\widehat{\xi}^{\log}(x)\| < 10^{-6}$, while the Hausdorff errors are of order between $10^{-1}$ and $10^{-2}$ (see Figure \ref{fig:consistency}), the numerical errors induced by the SCMS algorithm are dominated by the statistical error and can be safely ignored. 

As illustrated in Figure \ref{fig:consistency}, the Hausdorff errors exhibit an approximate linear decay on the log-log scale. As presented in Theorem \ref{thm:scms_total_convergence} and Remark \ref{rem:original_stopping_time}, theoretically statistical error is asymptotically bounded by a rate of $O(n^{-2/(d+8)}) = O(n^{-0.2})$ up to a logarithmic factor, given $d=2$ and our bandwidth selection rule. We fitted a linear regression based on the logarithm of the mean Hausdorff error and the logarithm of the sample sizes across all simulated scenarios, yielding a fitted regression slope of $-0.326$, which is faster than the theoretical upper bound of $-0.2$. This occurs because our simulation represents a special scenario where the density is entirely flat along the ridge, making the ridge a subset of critical points where $\nabla f=0$. Hence, the convergence enjoys a faster rate than the one we develop here for a more generic setting, because the rate of convergence for gradient estimation is faster than that for Hessian estimation using KDE. Also see \cite{qiao2025confidence}, where both scenarios are discussed for static ridge estimation.

\section{Discussion}
\label{sec:conclusion}

In this paper, we resolved an important open problem in the literature surrounding Subspace Constrained Mean Shift (SCMS) algorithms by introducing a paradigm shift from static density ridges to stable density ridges. We proved that the SCMS algorithm estimates the latter, which is defined by the Jacobian of the vector field associated with the SCMS algorithm and its corresponding eigenvalues. By formalizing the ridge-regular function class, we established the uniform R-linear convergence and topological surjectivity of both the continuous flow and the discrete algorithmic maps.

Furthermore, our joint analysis of statistical and computational errors exposed a computational bottleneck in the original SCMS algorithm: coupling the step size to the smoothing bandwidth forces the iteration complexity to blow up polynomially as the sample size increases. By decoupling the step size from the bandwidth, we showed that practitioners can recover the log-density ridge in $O(\log n)$ iterations, making the generalized SCMS algorithm highly feasible for massive datasets.

Currently, our theoretical framework is localized, focusing on the convergence and surjectivity of trajectories initiated within a compact neighborhood of the stable ridge. An open question is characterizing the behavior of the continuous vector field $\xi$ and the SCMS algorithms beyond this neighborhood, or more broadly, over the entire space $\mathbb{R}^d$. As our numerical tests in Section \ref{sec:numerical_experiments} have demonstrated, global convergence from far outside this neighborhood is empirically viable. Naturally, one can extend the SCMS flow backwards in time to define the basins of attraction for the stable ridge. In classical Morse theory~\citep{milnor1963morse}, the stable manifolds (basins of attraction) of local maxima for a Morse function partition the domain, covering the entire space up to a set of Lebesgue measure zero corresponding to the boundaries between adjacent basins. See, e.g., \cite{arias2025clustering}, for the analysis of hill-climbing (including Mean Shift) algorithms for the purpose of space partitioning (or clustering).

Whether the attraction regions of stable ridges under the SCMS flow similarly partition $\mathbb{R}^d$ almost everywhere remains to be investigated. Characterizing the geometric and topological conditions on the density function $f$ under which this global partitioning occurs represents an interesting direction for future research.

\bibliographystyle{chicago}
\bibliography{library}

\appendix
\section{Static vs. Stable Ridges}
\label{app:topological_divergence}

As discussed in the Introduction section, the local concavity of the density ($\lambda_{k+1} < 0$) and the dynamic stability of the continuous flow ($\mathrm{Re}(\mu_{k+1}) < 0$) are distinct conditions for the static ridge and stable ridge, respectively. Because the Jacobian $J_\xi$ incorporates the rotation of the trailing eigenspace, these two conditions only coincide under highly regular curvature conditions. In general, the static ridge and stable ridge are different structures. 

To quantify this distinction for a 1-dimensional ridge ($k=1$) in $\mathbb{R}^2$, we derive the eigenvalues and eigenvectors of the Jacobian at a point $x$ where $\xi(x)=0$. By the product rule, the Jacobian of the vector field $\xi(x) = \Pi(x)\nabla f(x)$ is 
\begin{align}
\label{Jacobian Ex}
J_\xi(x) = \Pi(x) \nabla^2 f(x) + (\nabla \Pi(x)) \nabla f(x).
\end{align}
Let $v_1(x)$ and $v_2(x)$ be the unit eigenvectors corresponding to the leading and trailing Hessian eigenvalues ($\lambda_1(x) \ge \lambda_2(x)$), so the projection matrix is $\Pi(x) = v_2(x) v_2(x)^\top$. 

Below we verify that $v_2$ is an eigenvector of $J_\xi(x)$. To show this, we multiply both sides of \eqref{Jacobian Ex} by $v_2$. Since $\xi(x) = 0$, the first term becomes $\Pi(x) \nabla^2 f(x) v_2(x) = \Pi(x) (\lambda_2(x) v_2(x)) = \lambda_2(x) v_2(x)$. The second term becomes
\begin{equation*}
(\nabla_{v_2} \Pi) \nabla f = (\nabla_{v_2} v_2) (v_2^\top \nabla f) + v_2 (\nabla_{v_2} v_2)^\top \nabla f = v_2 \langle \nabla_{v_2} v_2, \nabla f \rangle.
\end{equation*}
Letting $\Delta_{\text{rot}} = \langle \nabla_{v_2} v_2, \nabla f \rangle$, we can write
\begin{equation}\label{eq:jacobian_eigenvector_app}
J_\xi(x) v_2 = \big( \lambda_2 + \Delta_{\text{rot}} \big) v_2.
\end{equation}

Equation \eqref{eq:jacobian_eigenvector_app} proves that $v_2$ is an eigenvector of the Jacobian. To determine its index in the sorted eigenvalues $\mathrm{Re}(\mu_1) \ge \mathrm{Re}(\mu_2)$, we multiply both sides of \eqref{Jacobian Ex} by $v_1$. For the first term, $\Pi(x) \nabla^2 f(x) v_1 = \Pi(x)(\lambda_1 v_1) = 0$, because $v_1$ is in the null space of the projection matrix $\Pi(x)$. Following the same derivation for the second term, we find $J_\xi(x) v_1 = c v_2$, where $c = \langle \nabla_{v_1} v_2, \nabla f \rangle$. 

Since $J_\xi(x)v_1 = 0v_1 + c v_2$ and $J_\xi(x)v_2 = 0v_1 + (\lambda_2 + \Delta_{\text{rot}})v_2$, the Jacobian matrix $J_\xi(x)$ expressed in the $(v_1, v_2)$ coordinate basis takes the lower-triangular form
\begin{equation}
\begin{bmatrix} 0 & 0 \\ c & \lambda_2 + \Delta_{\text{rot}} \end{bmatrix}.
\end{equation}
Hence the eigenvalues of $J_\xi(x)$ are $\{0, \lambda_2 + \Delta_{\text{rot}}\}$, which are the diagonal entries of the triangular matrix. According to our sorting rule $\mathrm{Re}(\mu_1) \ge \mathrm{Re}(\mu_2)$, we can write $\mu_1 = \max(0, \lambda_2 + \Delta_{\text{rot}})$ and $\mu_2 = \min(0, \lambda_2 + \Delta_{\text{rot}})$. By the definition of the stable ridge in \eqref{eq:stable_ridge}, a point $x$ such that $\xi(x)=0$ belongs to the stable ridge if and only if $\lambda_2(x) + \Delta_{\text{rot}}(x) < 0$, where the term $\Delta_{\text{rot}}$ quantifies how the trailing eigenspace rotates as probability mass flows toward the ridge.

\subsection{Examples of Ridge Divergence and Coincidence}
\label{app:topological_examples}

\begin{figure}[htbp]
    \centering
    \includegraphics[width=\textwidth]{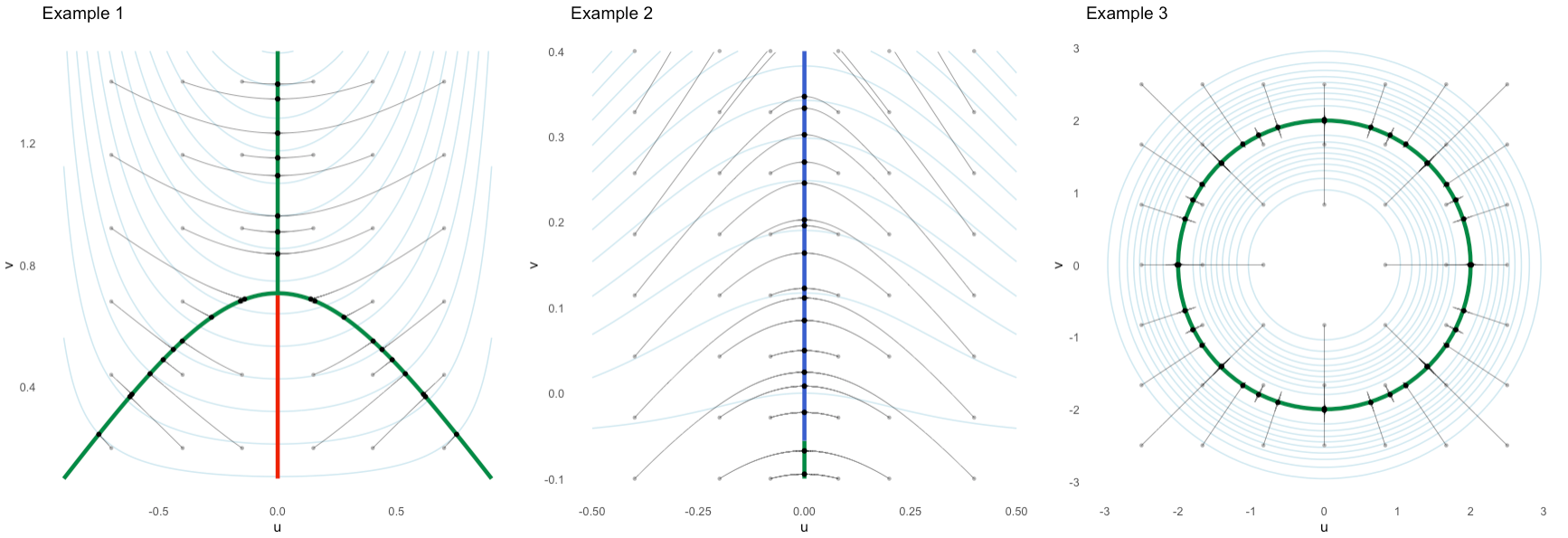}
    \caption{Three examples illustrating the difference and coincidence between static and stable ridges. Green segments indicate coincidence ($\mathcal{R}_{\text{stable}} \cap \mathcal{R}_{\text{static}}$), red indicates static-only branches that are dynamically unstable ($\mathcal{R}_{\text{static}} \setminus \mathcal{R}_{\text{stable}}$), and blue indicates dynamic attractors located in static valleys ($\mathcal{R}_{\text{stable}} \setminus \mathcal{R}_{\text{static}}$). Starting from grids of initial points (\textit{gray dots}), the SCMS continuous flows $\varphi_t$  (\textit{black lines}) converge to the limit points (\textit{black dots}) on the stable ridges, but not always on the static ridges.}
    \label{fig:topological_taxonomy}
\end{figure}

This subsection provides three examples comparing the static ridge ($\mathcal{R}_{\text{static}}$) and the stable ridge ($\mathcal{R}_{\text{stable}}$) for $k=1$ and $d=2$. As derived in the preceding derivation, the relation between static and stable ridges is determined by the sign of $\lambda_2 + \Delta_{\text{rot}}$. These examples are visualized in Figure \ref{fig:topological_taxonomy}. For each example, we only focus on the ridges restricted to a compact domain $\mathcal{D} \subset \mathbb{R}^2$. 

\subsubsection{Example 1: Qiao-Polonik Pitchfork Bifurcation}
\label{Pitchfork}

We first analyze the density $f(u, v) = C + \frac{3}{8}(1-u^2)v$ restricted to $\mathcal{D} = [-0.9, 0.9] \times [0.1, 1.5]$. This was introduced by \cite{qiao2025algorithms} as a counterexample to show that SCMS does not always converge to the static ridge. Here, and in the following examples, a positive constant $C > 0$ ensures $f$ is positive on $\mathcal{D}$ and is a valid probability density, and ridges remain invariant to a constant shift in the density. Let $c = 3/8$. The gradient and Hessian of $f$ are
\begin{equation*}
\nabla f(u,v) = c \begin{pmatrix} -2uv \\ 1-u^2 \end{pmatrix}, \quad  \nabla^2 f(u,v) = c \begin{pmatrix} -2v & -2u \\ -2u & 0 \end{pmatrix}.
\end{equation*}
The eigenvalues of the Hessian are $\lambda_{1,2} = c\left( -v \pm \sqrt{v^2 + 4u^2} \right)$. 

Both ridge definitions require $\xi(u,v) = 0$, which corresponds to the union of two sets: the vertical axis ($u=0$) and a pair of symmetric wings defined by $v = \psi(u) = \frac{1-u^2}{\sqrt{2(1+u^2)}}$. 

\textbf{The static ridge:} The condition for the static ridge is $\lambda_2 < 0$. On the vertical axis ($u=0$), the trailing eigenvalue is $\lambda_2 = -2cv$, which is negative for all $v > 0$. On the symmetric wings, the trailing eigenvalue is $\lambda_2 = c\left( -v - \sqrt{v^2 + 4u^2} \right)$, which is negative, since both $v$ and $c$ are positive across the domain. Therefore, the static ridge claims the entire pitchfork structure, including the vertical axis extending down to the bottom of the domain $\mathcal{D}$.

\textbf{The stable ridge:} We compute $\Delta_{\text{rot}}$. On the vertical axis, the trailing eigenvector is $v_2 = (1, 0)^\top$. It follows from straightforward calculations that $\nabla_{v_2} v_2 = (0, 1/v)^\top$ and $\nabla f = (0, c)^\top$, which gives $\Delta_{\text{rot}} = \frac{c}{v}$.
The dynamic stability condition on the axis requires
\begin{equation*}
\mu_2(0, v) = \lambda_2 + \Delta_{\text{rot}} = -2cv + \frac{c}{v} < 0 \implies v > \frac{1}{\sqrt{2}}.
\end{equation*}
Consequently, the lower axis below $1/\sqrt{2}$ loses dynamic stability and is not included in the stable ridge. Conversely, by calculation we have $\mu_2 = -c\sqrt{2(1+u^2)} \left[ 1 + \frac{(1-u^2)(3u^2-1)}{(3u^2+1)^2} \right]$ along the wings $v=\psi(u)$, which remains negative for all $u \neq 0$. Thus, the symmetric wings are retained in the stable ridge. The stable ridge drops the lower axis, defining a bifurcation at $v = 1/\sqrt{2}$. For this example, the stable ridge is a strict subset of the static ridge.

\subsubsection{Example 2: Static Valley as Stable Ridge}
\label{Valley Attractor}

Consider $\mathcal{D} = [-0.5, 0.5] \times [-0.1, 0.4]$ and a density defined as
\begin{equation*}
f(u, v) = C + v + 5v^2 + \frac{1}{2}(1 + 18v)u^2.
\end{equation*}
The derivatives are $\nabla f (u,v) = \begin{pmatrix} u(1+18v) \\ 1+10v+9u^2 \end{pmatrix}$ and $\nabla^2 f(u,v) = \begin{pmatrix} 1+18v & 18u \\ 18u & 10 \end{pmatrix}$.

Setting $\xi(u,v) = 0$ reveals that the vertical axis ($u=0$) is the only candidate set for both the static and stable ridges. On this axis, the Hessian is diagonal with $\nabla^2 f(0, v) = \text{diag}(1+18v, 10)$. Throughout $\mathcal{D}$ where $v \le 0.4$, the trailing eigenvalue is $\lambda_2 = 1+18v$, and the leading eigenvalue is $\lambda_1 = 10$.

\textbf{The static ridge:} The axis belongs to $\mathcal{R}_{\text{static}}$ only where $\lambda_2 < 0$. Within $\mathcal{D}$, this restricts the static ridge to the sub-interval $v \in [-0.1, -1/18)$. For any $v > -1/18$, we have $\lambda_2 > 0$. Geometrically, this positive curvature means the density is locally convex in the normal direction, making that portion of the vertical axis a static valley rather than a static ridge.

\textbf{The stable ridge:} We compute $\Delta_{\text{rot}}$. The trailing eigenvector is $v_2 = (1, 0)^\top$ and the leading eigenvector is $v_1 = (0, 1)^\top$. The directional derivative of the Hessian is $\nabla_{v_2} \nabla^2 f = \begin{pmatrix} 0 & 18 \\ 18 & 0 \end{pmatrix}$. By spectral perturbation theory, we have
\begin{equation*}
\nabla_{v_2} v_2 = \frac{v_1^\top (\nabla_{v_2} \nabla^2 f) v_2}{\lambda_2 - \lambda_1} v_1 = \frac{18}{(1+18v) - 10} \begin{pmatrix} 0 \\ 1 \end{pmatrix} = \begin{pmatrix} 0 \\ \frac{2}{2v-1} \end{pmatrix}.
\end{equation*}
Combining this with $\nabla f = (0, 1+10v)^\top$ yields $\Delta_{\text{rot}} = \frac{2(1+10v)}{2v-1}$ and
\begin{equation*}
\lambda_2 +  \Delta_{\text{rot}} = (1+18v) + \frac{2(1+10v)}{2v-1} = \frac{36v^2 + 4v + 1}{2v - 1}.
\end{equation*}
Because $36v^2 + 4v + 1>0$ for all $v$, the sign of $\lambda_2 +  \Delta_{\text{rot}}$ is determined by the denominator $2v-1$, which is negative for all $v \in [-0.1, 0.4]$, ensuring $\mu_2 = \lambda_2 +  \Delta_{\text{rot}} < 0$ on the entire vertical axis ($u=0$) within $\mathcal{D}$. The stable ridge spans the entire vertical line $v \in [-0.1, 0.4]$, even where it is a static valley for $v > -1/18$. For this example, the static ridge is a strict subset of the stable ridge.

\subsubsection{Example 3: Coincidence Between Static and Stable Ridges}
\label{Radial Symmetry}

We construct an example for which the two ridge definitions coincide, that is, $\mathcal{R}_{\text{stable}} = \mathcal{R}_{\text{static}}$. Consider a radially symmetric density $f(u,v) = g(\rho)$, where $\rho = \sqrt{u^2+v^2}$, formed by adding isotropic Gaussian noise to a uniform distribution on a 1-dimensional circle of radius $r$. We analyze a tubular neighborhood $\mathcal{D}$ where the profile $g(\rho)$ has a global maximum at $\rho = r^*$ and $g''(\rho) < 0$. Under the standard polar transformation $u = \rho \cos\theta$ and $v = \rho \sin\theta$, we define the orthonormal basis vectors $\mathbf{e}_\rho = (\cos\theta, \sin\theta)^\top$ and $\mathbf{e}_\theta = (-\sin\theta, \cos\theta)^\top$. (Note that while $\mathbf{e}_\rho$ indicates the direction of radial extension, its Cartesian orientation depends solely on the angle $\theta$).

The gradient is radial, that is, $\nabla f(u,v) = g'(\rho) \mathbf{e}_\rho$. By applying $\nabla \mathbf{e}_\rho = \frac{1}{\rho}\mathbf{e}_\theta \mathbf{e}_\theta^\top$, the Hessian is
\begin{equation*}
\nabla^2 f (u,v) = g''(\rho) \mathbf{e}_\rho \mathbf{e}_\rho^\top + \frac{g'(\rho)}{\rho} \mathbf{e}_\theta \mathbf{e}_\theta^\top.
\end{equation*}
The trailing eigenvector is $v_2 = \mathbf{e}_\rho$, with trailing eigenvalue $\lambda_2 = g''(\rho)$. The projected gradient is $\xi(u,v) = g'(\rho) \mathbf{e}_\rho$, which vanishes if and only if $g'(\rho) = 0$, occurring at $\rho = r^*$. 

\textbf{The static ridge:} Because $g''(r^*) < 0$, the static ridge is just the 1-dimensional circle of radius $r^*$.

\textbf{The stable ridge:} For this example, $\Delta_{\text{rot}} = 0$, because
\begin{equation*}
\nabla_{v_2} v_2 = (\nabla \mathbf{e}_\rho) \mathbf{e}_\rho = \left( \frac{1}{\rho}\mathbf{e}_\theta \mathbf{e}_\theta^\top \right) \mathbf{e}_\rho = \frac{1}{\rho} \mathbf{e}_\theta (\mathbf{e}_\theta^\top \mathbf{e}_\rho) = 0.
\end{equation*}
Consequently, $\mu_2 \equiv \lambda_2 = g''(r^*)$, confirming that the stable and static ridges are identical in this example.

\section{Proofs}
\label{app:proofs}

\subsection{Proofs for Section \ref{sec:population_geometry}}

\begin{myproof}[of Lemma \ref{lem:geometric_regularity}]
By Assumption (A1), $f \in C^4(\mathbb{R}^d, \mathbb{R})$. By Assumption (A2), the condition $\eta > 0$ ensures both $\Pi$ and $\xi$ are $C^2$-smooth on $\mathrm{int}(\mathcal{D})$. 

Consider a point $x \in \mathcal{R}(f)$. Let $V(x) \in \mathbb{R}^{d \times (d-k)}$ be a matrix whose columns form an orthogonal basis for the trailing eigenspace $E_\perp(x)$. By Property 2 of Definition \ref{def:function_class}, the symmetric part of the projected Jacobian $V(x)^\top J_\xi(x) V(x)$ is negative definite. Thus, $V(x)^\top J_\xi(x) V(x)$ is non-singular and has full rank $d-k$.

Let $U \subset \mathrm{int}(\mathcal{D})$ be a sufficiently small neighborhood around $x$. We construct a $C^2$-smooth map $\psi: U \to \mathbb{R}^{d-k}$ such that $\xi(y)=0$ is equivalent to $\psi(y)=0$ for all $y \in U$. Define $\widetilde{W}(y) = \Pi(y) V(x)$. Because $\Pi(x) V(x) = V(x)$ has full column rank $d-k$ and $\Pi$ is continuous, $\widetilde{W}(y)$ retains full column rank on $U$ for sufficiently small $U$. Applying Gram-Schmidt orthogonalization to the columns of $\widetilde{W}(y)$ yields an orthonormal basis for $E_\perp(y)$, denoted by the matrix $W(y) \in \mathbb{R}^{d \times (d-k)}$. Because the projection matrix $\Pi$ and the Gram-Schmidt operations are $C^2$-smooth, $W$ is $C^2$-smooth on $U$. Note that $W(x) = V(x)$ since the columns of $V(x)$ are already orthonormal.

By construction, $\Pi(y)=W(y)W(y)^\top$, yielding $\xi(y) = W(y)\psi(y)$, where
\begin{equation}
\psi(y) = W(y)^\top \nabla f(y).
\end{equation}
Since $W(y)$ has full column rank, the condition $\xi(y) = 0$ is equivalent to $\psi(y)=0$ for $y\in U$. Furthermore, $\psi$ is $C^2$-smooth on $U$ since both $W$ and $\nabla f$ are $C^2$-smooth. Let $\psi_i$ denote the $i$-th component of $\psi$, and $W_i$ denote the $i$-th column of $W$ for $i=1,\dots,d-k$. By the multivariate product rule, the Jacobian of $\xi(y) = W(y)\psi(y)$ evaluated at $x$ is
\begin{equation}
J_\xi(x) = W(x) J_\psi(x) + \sum_{i=1}^{d-k} \psi_i(x) \nabla W_i(x) = V(x) J_\psi(x),
\end{equation}
because $\psi(x) = 0$. Left-multiplying by $V(x)^\top$ yields $V(x)^\top J_\xi(x) = J_\psi(x)$. 

Since $V(x)^\top J_\xi(x) V(x)$ is non-singular, $J_\psi(x)$ has full row rank $d-k$. By the Regular Level Set Theorem \citep[Corollary 5.14]{lee2012introduction}, $\psi^{-1}(0)$ forms a $C^2$-smooth surface of dimension $k$ within $U$. Because $x \in \mathcal{R}(f)$ is arbitrary, $\mathcal{R}(f)$ is a global $C^2$-smooth $k$-dimensional submanifold.

Fix an $\epsilon \in (0, \kappa]$. By Property 3 of Definition \ref{def:function_class}, any point $z \in \mathcal{D}$ satisfying $\|\xi(z)\| \le \epsilon \le \kappa$ must satisfy $\mathrm{Re}(\mu_{k+1}(z)) \le -\gamma/2$. Consequently, the set can be equivalently expressed as
\begin{equation}
\label{Repsilon}
\mathcal{R}_{\epsilon}(f) = \left\{ z \in \mathcal{D} : \|\xi(z)\| \le \epsilon \text{ and } \mathrm{Re}(\mu_{k+1}(z)) \le -\frac{\gamma}{2} \right\}.
\end{equation}
Since $\xi$ and $\mathrm{Re}(\mu_{k+1})$ are continuous functions, $\mathcal{R}_{\epsilon}(f)$ is a closed subset of the compact domain $\mathcal{D}$, making it compact. Property 2 dictates $\mathcal{R}_{\epsilon}(f) \subset \mathrm{int}(\mathcal{D})$. Thus, $\mathcal{R}(f) = \xi^{-1}(0) \cap \mathcal{R}_{\epsilon}(f)$ is the intersection of a closed set and a compact set, which is compact. Finally, because $\mathcal{R}(f) \subset \mathrm{int}(\mathcal{D})$, it has no boundary.
\end{myproof}

\begin{myproof}[of Lemma \ref{lem:continuous_limit}]
Let $x \in \mathcal{R}_{\epsilon}(f)$. Using the definition in \eqref{varphi_def}, we have
\begin{align}
\frac{d}{dt} \|\xi(\varphi_t(x))\|^2 &= 2 \xi(\varphi_t(x))^\top \left( \frac{d}{dt} \xi(\varphi_t(x)) \right) \\
&= 2 \xi(\varphi_t(x))^\top J_\xi(\varphi_t(x)) \xi(\varphi_t(x)).
\end{align}
Since $\xi(\varphi_t(x)) \in E_\perp(\varphi_t(x))$, Property 2 of Definition \ref{def:function_class} implies
\begin{equation}\label{eq:time_deriv_speed}
\frac{d}{dt} \|\xi(\varphi_t(x))\|^2 \le -2\gamma \|\xi(\varphi_t(x))\|^2.
\end{equation}
By Gr\"{o}nwall's inequality, \eqref{eq:time_deriv_speed} implies
\begin{equation}\label{eq:gronwall_decay}
\|\xi(\varphi_t(x))\|^2 \le e^{-2\gamma t} \|\xi(x)\|^2 \implies \|\xi(\varphi_t(x))\| = \|\dot{\varphi}_t(x)\| \le e^{-\gamma t} \|\xi(x)\|.
\end{equation}
Since $e^{-\gamma t} \le 1$ for $t \ge 0$, we have $\|\xi(\varphi_t(x))\| \le \|\xi(x)\| \le \epsilon$. Furthermore, because $\epsilon \le \kappa$, Property 3 of Definition \ref{def:function_class} guarantees $\mathrm{Re}(\mu_{k+1}(\varphi_t(x))) \le -\gamma/2 < 0$. Therefore, $\varphi_t(x) \in \mathcal{R}_{\epsilon}(f)$ for all $t \ge 0$.

The total arc length $\mathcal{L}(x)$ of $\varphi_t(x)$ satisfies
\begin{equation}
\mathcal{L}(x) = \int_0^\infty \|\dot{\varphi}_t(x)\| \, dt \le \int_0^\infty e^{-\gamma t} \|\xi(x)\| \, dt = \frac{1}{\gamma} \|\xi(x)\|.
\end{equation}

Since the integral on the left-hand side converges, the trajectory $\varphi_t(x)$ satisfies the Cauchy criterion as $t \to \infty$ in $\mathbb{R}^d$. Thus, $\varphi_t(x)$ converges to a unique limit point $\Phi(x) = \lim_{t \to \infty} \varphi_t(x)$. 

By \eqref{eq:gronwall_decay}, $\lim_{t \to \infty} \|\xi(\varphi_t(x))\| = 0$. By the continuity of $\xi$ on $\mathcal{D}$, $\|\xi(\Phi(x))\| = 0$. Because $\|\xi(\Phi(x))\| = 0 < \kappa$, Property 3 of Definition \ref{def:function_class} implies $\mathrm{Re}(\mu_{k+1}(\Phi(x))) \le -\frac{\gamma}{2} < 0$. Thus, $\Phi(x) \in \mathcal{R}(f)$.
\end{myproof}

of
\begin{myproof}[of Corollary \ref{cor:hausdorff_bound}]
Since $\mathcal{R}(f) \subseteq \mathcal{R}_\epsilon(f)$, we have $d(\mathcal{R}(f) | \mathcal{R}_\epsilon(f)) = 0$, and thus $d_H(\mathcal{R}_\epsilon(f), \mathcal{R}(f)) = d(\mathcal{R}_\epsilon(f) | \mathcal{R}(f))$. 

Let $x \in \mathcal{R}_\epsilon(f)$. By Lemma \ref{lem:continuous_limit}, we have $\Phi(x) \in \mathcal{R}(f)$, which yields
\begin{equation}
\label{hausdorff_upper_bound}
    \inf_{z \in \mathcal{R}(f)} \|x - z\| \le \|x - \Phi(x)\| \le \mathcal{L}(x) \le \frac{\|\xi(x)\|}{\gamma} \le \frac{\epsilon}{\gamma}.
\end{equation}
Taking the supremum over $x \in \mathcal{R}_\epsilon(f)$ establishes the upper bound in \eqref{hausdorff_bound}. 

Next we consider the lower bound. Let $\tilde{x} \in \mathcal{R}_\epsilon(f)$ such that $\|\xi(\tilde{x})\| = \epsilon$. Because $\mathcal{R}(f)$ is compact by Lemma \ref{lem:geometric_regularity}, there exists $x_* \in \mathcal{R}(f)$ such that $\inf_{z \in \mathcal{R}(f)} \|\tilde{x} - z\| = \|\tilde{x} - x_*\|$. Denote $v = \tilde{x} - x_*$.

By \eqref{hausdorff_upper_bound}, we have $\|v\| \le \epsilon/\gamma$. We first show the Euclidean ball $B_{\epsilon/\gamma}(x_*)$ is contained in $\mathcal{R}_{\kappa}(f)$. For any $y \in B_{\epsilon/\gamma}(x_*)$, define the line segment $z(t) = x_* + t(y - x_*)$ for $t \in [0,1]$. Because $x_* \in \mathcal{R}(f)$ is in the interior of $\mathcal{R}_{\kappa}(f)$, the segment begins in $\mathcal{D}$. As long as the partial segment $z([0,t]) \subseteq \mathcal{D}$, the mean value theorem ensures $\|\xi(z(t))\| \le L \|z(t) - x_*\| \le L (\epsilon/\gamma) \le \kappa$, due to $\epsilon \le \frac{\gamma}{L} \kappa$. This implies the entire segment $z([0,1])$ cannot exit $\mathcal{R}_{\kappa}(f) \subseteq \mathcal{D}$, guaranteeing $B_{\epsilon/\gamma}(x_*)$ is contained in $\mathcal{D}$. 

Because $B_{\epsilon/\gamma}(x_*)$ is convex, the line segment connecting $x_*$ and $\tilde{x}$ lies entirely within $\mathcal{D}$. Since $\xi \in C^2(\mathcal{D})$ and $\xi(x_*) = 0$, a Taylor expansion yields
\begin{equation}
    \xi(\tilde{x}) = \xi(x_*) + J_\xi(x_*)v + R(v) = J_\xi(x_*)v + R(v),
\end{equation}
where $R(v)$ is the second-order remainder term. By the triangle inequality, we can write
\begin{equation}\label{eq:taylor_boundary}
    \|\xi(\tilde{x})\| = \|J_\xi(x_*) v + R(v)\| \le \|J_\xi(x_*) v\| + \|R(v)\| \le L \|v\| + M \|v\|^2,
\end{equation}
where in the last inequality we have used the definition $L = \sup_{x \in \mathcal{D}} \|J_\xi(x)\|$ and $M = \frac{1}{2} \sup_{x \in \mathcal{D}} \|\nabla^2 \xi(x)\|$. Substituting $\|\xi(\tilde{x})\| = \epsilon$ into \eqref{eq:taylor_boundary} yields
\begin{equation}
    \epsilon \le L \|v\| + M \|v\|^2 \le (L + M\epsilon/\gamma) \|v\|,
\end{equation}
where we use $\|v\| \le \epsilon/\gamma$ from \eqref{hausdorff_upper_bound}. Because $\epsilon \le \frac{L\gamma}{M}$, it follows that
\begin{equation}
  \inf_{z \in \mathcal{R}(f)} \|\tilde{x} - z\| = \|v\| \ge \frac{1}{L + M\epsilon/\gamma} \epsilon \ge \frac{1}{2L} \epsilon,
\end{equation}
establishing the lower bound in \eqref{hausdorff_bound}.
\end{myproof}

\begin{myproof}[of Theorem \ref{thm:continuous_surjectivity}]
Let $x_* \in \mathcal{R}(f)$. By definition, $\xi(x_*) = 0$, making $x_*$ an equilibrium point of $\xi$. Because $\mathcal{R}(f)$ is a $k$-dimensional submanifold of $\mathcal{D}$ (Lemma \ref{lem:geometric_regularity}), $J_\xi(x_*)$ possesses $k$ zero eigenvalues corresponding to the tangent space. By Property 2 of Definition \ref{def:function_class}, the remaining $d-k$ eigenvalues have real parts upper bounded by $-\gamma < 0$. The Center Manifold Theorem \citep[Sec 2.7]{perko2001differential} guarantees the existence of a $C^2$-smooth, $(d-k)$-dimensional manifold $W^s_{\text{loc}}(x_*)$ locally passing through $x_*$, defined as
\begin{equation}
    W^s_{\text{loc}}(x_*) = \big\{ y \in U:\; \lim_{t \to \infty} \varphi_t(y) = x_* \big\},
\end{equation}
where $U$ is a neighborhood of $x_*$. By \eqref{Phi def}, we have $\Phi(y) = x_*$ for all $y \in W^s_{\text{loc}}(x_*)$.

Let $g(x) = \|\xi(x)\|$. Because $k < d$, we can choose $y \in W^s_{\text{loc}}(x_*) \setminus \{x_*\}$ such that $0 < g(y) < \epsilon$. Consider the trajectory $z(t)$ governed by the reversed flow $\dot{z}(t) = -\xi(z(t))$ with initial condition $z(0) = y$. Following the derivation in Lemma \ref{lem:continuous_limit}, while $z(t) \in \mathcal{R}_{\epsilon}(f)$, we have
\begin{equation}
    \frac{d}{dt} \|\xi(z(t))\|^2 = -2 \xi(z(t))^\top J_\xi(z(t)) \xi(z(t)) \ge 2\gamma \|\xi(z(t))\|^2 > 0.
\end{equation}
By Gr\"{o}nwall's inequality, we obtain $\|\xi(z(t))\|^2 \ge e^{2\gamma t}\|\xi(y)\|^2$. Because $\|\xi(z(t))\|$ grows exponentially, $z(t)$ cannot remain in the bounded sub-level set $\{x \in \mathcal{D}: g(x) < \epsilon\}$ for all $t \ge 0$. Since $\mathcal{R}_{\epsilon}(f)$ is a compact set contained in $\mathrm{int}(\mathcal{D})$ (see \eqref{Repsilon}), the continuous trajectory $z(t)$ must intersect the boundary $\partial \mathcal{R}_{\epsilon}(f)$. By the Intermediate Value Theorem, there exists a first exit time $t_0 > 0$ such that $g(z(t_0)) = \epsilon$. 

For all $t \in [0, t_0]$, we have $g(z(t)) \le \epsilon \le \kappa$. By Property 3 of Definition \ref{def:function_class}, this ensures $\mathrm{Re}(\mu_{k+1}(z(t))) \le -\gamma/2 < 0$, which validates the domain of the expansion bound up to $t_0$.

Let $x_0 = z(t_0)$. Because $g(x_0) = \epsilon$ and $\mathrm{Re}(\mu_{k+1}(x_0)) \le -\gamma/2 < 0$, it follows that $x_0 \in \partial \mathcal{R}_{\epsilon}(f)$. Because $x_0$ is reached by the reversed flow $z(t)$ from $y$, the forward flow $\varphi_t(x_0)$ passes through $y \in W^s_{\text{loc}}(x_*)$. Thus, $\lim_{t \to \infty} \varphi_t(x_0) = \lim_{t \to \infty} \varphi_t(y) = x_*$, which yields $\Phi(x_0) = x_*$. Since $x_* \in \mathcal{R}(f)$ is arbitrary, $\Phi$ is surjective.
\end{myproof}

\begin{myproof}[of Lemma \ref{lem:lipschitz_projection}]
As stated in the beginning of the proof of Lemma \ref{lem:geometric_regularity}, $\xi$ is $C^2$-smooth on $\mathrm{int}(\mathcal{D})$. This implies that for any finite time $t \ge 0$, the flow $\varphi_t(x)$ is continuously differentiable with respect to its initial condition $x\in \mathcal{R}_{\kappa}(f) \subset \mathrm{int}(\mathcal{D})$. 

Let $Y_x(t) = J_{\varphi_t}(x)$. Differentiating $\dot{\varphi}_t(x) = \xi(\varphi_t(x))$ with respect to $x$ yields the variational equation
$$\dot{Y}_x(t) = J_\xi(\varphi_t(x)) Y_x(t), \quad \text{with } Y_x(0) = I.$$
Let $z = \Phi(x) \in \mathcal{R}(f)$. By defining $A_z = J_\xi(z)$ and $B_z(t) = J_\xi(\varphi_t(x)) - J_\xi(z)$, we can write
$$\dot{Y}_x(t) = \left[ A_z + B_z(t) \right] Y_x(t).$$
Applying the variation of constants formula yields
\begin{align}
\label{Yx formula}
Y_x(t) = e^{A_z t} + \int_0^t e^{A_z (t-s)} B_z(s) Y_x(s) \, ds.
\end{align}

By Property 2 of Definition \ref{def:function_class}, the spectrum of $A_z$ consists of an eigenvalue $0$ of algebraic multiplicity $k$, and $d-k$ eigenvalues with real parts bounded above by $-\gamma < 0$. Because $\mathcal{R}(f)$ is a $k$-dimensional manifold of equilibria, the tangent space $T_z\mathcal{R}(f)$ is contained in the null space of $A_z$. Thus, the geometric multiplicity of the zero eigenvalue equals $k$. Then by the real Schur decomposition, there exists an orthogonal matrix $U_z = [U_{1z}, U_{2z}]$ such that the $k$ columns of $U_{1z}$ form an orthonormal basis for the null space of $A_z$. Since $A_z U_{1z} = 0$, we obtain
$$A_z = U_z (U_z^\top A_z U_z) U_z^\top = U_z \begin{bmatrix} 0_{k \times k} & A_{1z} \\ 0_{(d-k)\times k} & A_{2z} \end{bmatrix} U_z^\top,$$
where $A_{1z} = U_{1z}^\top A_z U_{2z}$ and $A_{2z} = U_{2z}^\top A_z U_{2z}$. Here $A_{2z} \in \mathbb{R}^{(d-k) \times (d-k)}$ is a quasi-upper-triangular matrix whose $1 \times 1$ and $2 \times 2$ diagonal blocks correspond to the non-zero eigenvalues of $A_z$, all of which have real parts bounded above by $-\gamma$.

Define $E_z(t) = U_z^\top e^{A_z t} U_z$, which satisfies $\dot{E}_z(t) = (U_z^\top A_z U_z) E_z(t)$ with $E_z(0) = I_{k \times k}$. Partitioning $E_z(t)$ into $2 \times 2$ blocks of the same sizes as $A_z$ yields
$$\dot{E}_z(t) = \begin{bmatrix} \dot{E}_{1z}'(t) & \dot{E}_{1z}(t) \\ \dot{E}_{2z}'(t) & \dot{E}_{2z}(t) \end{bmatrix} = \begin{bmatrix} 0_{k \times k} & A_{1z} \\ 0_{(d-k)\times k} & A_{2z} \end{bmatrix} \begin{bmatrix} E_{1z}'(t) & E_{1z}(t) \\ E_{2z}'(t) & E_{2z}(t) \end{bmatrix}.$$
It can be seen that $E_{2z}'(t) = 0$, $E_{1z}'(t) = I_{k \times k}$, and $E_{2z}(t) = e^{A_{2z}t}$. Subsequently $\dot{E}_{1z}(t) = A_{1z} e^{A_{2z}t}$, that is,
\begin{align}
\label{E1z}
E_{1z}(t) = A_{1z} \int_0^t e^{A_{2z} s} \, ds = A_{1z} A_{2z}^{-1} (e^{A_{2z} t} - I) = A_{1z} A_{2z}^{-1} (E_{2z}(t) - I).
\end{align}

Recall that $L = \sup_{y \in \mathcal{D}} \|J_\xi(y)\|$. Since $U_z$ is orthogonal, we have
\begin{align}
\label{A12z bound}
\max(\|A_{1z}\|, \|A_{2z}\|) \le \|A_z\| = \|J_\xi(z)\| \le L  < \infty,
\end{align}
for all $z \in \mathcal{R}(f)$. Define the set of matrices
$$\mathcal{W} = \Big\{ W \in \mathbb{R}^{(d-k) \times (d-k)} : \|W\| \le L \text{ and } \max_j \mathrm{Re}(\lambda_j(W)) \le -\gamma \Big\}.$$
By continuity of the matrix 2-norm and eigenvalues, $\mathcal{W}$ is compact. By definition, $A_{2z} \in \mathcal{W}$ for all $z \in \mathcal{R}(f)$. Define $\psi: \mathcal{W} \times [0, \infty) \to \mathbb{R}$ by
$$\psi(W, t) = \|e^{Wt}\| e^{\gamma t / 2} = \|e^{(W + (\gamma/2)I)t}\|.$$
For $W \in \mathcal{W}$, the eigenvalues of $W + (\gamma/2)I$ have real parts bounded above by $-\gamma/2 < 0$. This ensures $\lim_{t \to \infty} \psi(W, t) = 0$. Then since $\psi$ is continuous on $\mathcal{W}$, $\psi$ attains a maximum on $\mathcal{W} \times [0, \infty)$. Let $C_{\max} = \max(1, \sup_{(W, t)} \psi(W, t)) < \infty$. Then for all $z \in \mathcal{R}(f)$,
\begin{align}
\label{E2z bound}
\|E_{2z}(t)\| = \psi(A_{2z}, t) e^{-\gamma t / 2} \le C_{\max} e^{-\gamma t / 2}.
\end{align}
Using \eqref{E1z}, it follows that
\begin{align}
\label{E1z bound}
\|E_{1z}(t)\| \le \|A_{1z}\| \|A_{2z}^{-1}\| (\|E_{2z}(t)\| + 1) \le \|A_{1z}\| \|A_{2z}^{-1}\| (C_{\max} + 1).
\end{align}

Because $e^{A_z t} = U_z E_z(t) U_z^\top$ and $U_z$ is orthogonal, 
\begin{align}
\label{eAz bound}
\|e^{A_z t}\| =\|E_z(t)\| = \left\| \begin{bmatrix} I_{k \times k} & E_{1z}(t) \\ 0_{(d-k) \times k} & E_{2z}(t) \end{bmatrix} \right\| \le 1 + \|E_{1z}(t)\| + \|E_{2z}(t)\|.
\end{align}
By the definition of $\mathcal{W}$, $0$ is not an eigenvalue of any $W \in \mathcal{W}$, and hence every matrix in $\mathcal{W}$ is invertible. Since the map $W \mapsto \|W^{-1}\|$ is continuous on the compact set $\mathcal{W}$, we have $K_\gamma := \max_{W \in \mathcal{W}} \|W^{-1}\| < \infty.$ Therefore, $\|A_{2z}^{-1}\| \le K_\gamma$ for all $z \in \mathcal{R}(f)$. 
It then follows from \eqref{A12z bound}, \eqref{E2z bound}, \eqref{E1z bound}, and \eqref{eAz bound} that
\begin{align}
\label{eAzt bound}
\|e^{A_z t}\| \le 1 + L K_\gamma (C_{\max} + 1) + C_{\max} = :\kappa_{\max} <\infty.
\end{align}

Recall that $M = \frac{1}{2} \sup_{x \in \mathcal{D}} \|\nabla^2 \xi(x)\| < \infty$. Since $\lim_{\tau \to \infty} \varphi_\tau(x) = z$, we can write
$$B_z(t) = J_\xi(\varphi_t(x)) - J_\xi(z) = -\int_t^\infty \nabla^2 \xi(\varphi_\tau(x)) \xi(\varphi_\tau(x)) \, d\tau.$$
By Lemma \ref{lem:continuous_limit}, $\|\xi(\varphi_\tau(x))\| \le e^{-\gamma \tau} \|\xi(x)\| \le \kappa e^{-\gamma \tau}$. As a result,
\begin{align}
\label{Bzt bound}
\|B_z(t)\| \le \int_t^\infty 2M \kappa e^{-\gamma \tau} \, d\tau = \frac{2M \kappa}{\gamma} e^{-\gamma t}.
\end{align}
Plugging the bounds in \eqref{eAzt bound} and \eqref{Bzt bound} into \eqref{Yx formula}, we have
$$\|Y_x(t)\| \le \kappa_{\max} + \int_0^t \kappa_{\max} \left( \frac{2M \kappa}{\gamma} e^{-\gamma s} \right) \|Y_x(s)\| \, ds.$$
By Gr\"{o}nwall's inequality, we obtain
$$\|Y_x(t)\| \le \kappa_{\max} \exp\left( \int_0^t \frac{2\kappa_{\max} M \kappa}{\gamma} e^{-\gamma s} \, ds \right) \le \kappa_{\max} \exp\left( \frac{2 \kappa_{\max} M \kappa}{\gamma^2} \right) =: L_\Phi < \infty.$$
It follows from \eqref{E2z bound} that $\lim_{t \to \infty} E_{2z}(t) = 0$, which, via \eqref{E1z}, further leads to $$\lim_{t \to \infty} E_{1z}(t) = \lim_{t \to \infty} A_{1z} A_{2z}^{-1} (E_{2z}(t) - I) = -A_{1z} A_{2z}^{-1},$$ and $$D_z(\infty) := \lim_{t \to \infty} e^{A_z t} =\lim_{t \to \infty} U_z E_z(t) U_z^\top =U_z \begin{bmatrix} I_{k \times k} & -A_{1z} A_{2z}^{-1} \\ 0 & 0 \end{bmatrix} U_z^\top.$$
In \eqref{Yx formula}, for the integral term, because $\|e^{A_z(t-s)}\| \le \kappa_{\max}$ for all $t \ge s$ and $\|B_z(s) Y_x(s)\|$ decays exponentially, the integrand $e^{A_z(t-s)} B_z(s) Y_x(s)$ is pointwise convergent as $t \to \infty$ and is uniformly dominated by an integrable function. By the Dominated Convergence Theorem, the limit $Y_x(\infty) = \lim_{t \to \infty} Y_x(t)$ exists in the form of $$Y_x(\infty) = D_z(\infty) + \int_0^\infty D_z(\infty) B_z(s) Y_x(s) \, ds.$$
and satisfies $\|Y_x(\infty)\| \le L_\Phi$. In fact, we can establish uniform convergence of $Y_x(t)$ as $t \to \infty$ over $x \in \mathcal{R}_{\kappa}(f)$. Notice that
\begin{align}
\label{Yxdiff}
Y_x(t) - Y_x(\infty) = (e^{A_z t} - D_z(\infty)) + \int_0^t (e^{A_z (t-s)} - D_z(\infty)) B_z(s) Y_x(s) \, ds - \int_t^\infty D_z(\infty) B_z(s) Y_x(s) \, ds.
\end{align}
For any $\tau \ge 0$, by using \eqref{E1z}, we have 
$$e^{A_z \tau} - D_z(\infty) = U_z \begin{bmatrix} 0 & E_{1z}(\tau) + A_{1z}A_{2z}^{-1} \\ 0 & E_{2z}(\tau) \end{bmatrix} U_z^\top = U_z \begin{bmatrix} 0 & A_{1z} A_{2z}^{-1} E_{2z}(\tau) \\ 0 & E_{2z}(\tau) \end{bmatrix} U_z^\top,$$
and subsequently by \eqref{A12z bound}, \eqref{E2z bound} and \eqref{E1z bound},
$$\|e^{A_z \tau} - D_z(\infty)\| \le (1 + \|A_{1z}\| \|A_{2z}^{-1}\|) \|E_{2z}(\tau)\| \le (1 + L K_\gamma) C_{\max} e^{-\gamma \tau / 2} \le \kappa_{\max} e^{-\gamma \tau / 2}.$$
where we have used \eqref{eAzt bound}, and by using it again we have $\|D_z(\infty)\| \le 2 \kappa_{\max}$. Plugging these established bounds as well as \eqref{Bzt bound} and $\|Y_x(s)\| \le L_\Phi$ into \eqref{Yxdiff} yields
\begin{align*}
& \|Y_x(t) - Y_x(\infty)\| \\
& \le \kappa_{\max} e^{-\gamma t / 2} + \int_0^t \kappa_{\max} e^{-\gamma(t-s)/2} \left(\frac{2M \kappa L_\Phi}{\gamma}\right) e^{-\gamma s} \, ds + 2\int_t^\infty \kappa_{\max} \left(\frac{2M \kappa L_\Phi}{\gamma}\right) e^{-\gamma s} \, ds \\
& = \kappa_{\max} e^{-\gamma t / 2} + \kappa_{\max} \left(\frac{2M \kappa L_\Phi}{\gamma}\right) e^{-\gamma t / 2} \int_0^t e^{-\gamma s / 2} \, ds + 2\kappa_{\max} \left(\frac{2M \kappa L_\Phi}{\gamma^2}\right) e^{-\gamma t} \\
& \le \kappa_{\max} e^{-\gamma t / 2} + \kappa_{\max} \left(\frac{8M \kappa L_\Phi}{\gamma^2}\right) e^{-\gamma t / 2}.
\end{align*}

This confirms that $Y_x(t)$ converges uniformly to $Y_x(\infty)$ over $\mathcal{R}_{\kappa}(f)$. Because $\varphi_t(x)$ converges pointwise to $\Phi(x)$ and $J_{\varphi_t}(x) = Y_x(t)$ converges uniformly to $Y_x(\infty)$, $\Phi(x)$ is continuously differentiable with $J_\Phi(x) = Y_x(\infty)$. Thus, $\sup_{x \in \mathcal{R}_{\kappa}(f)} \|J_\Phi(x)\| \le L_\Phi$.
\end{myproof}

\subsection{Proofs for Section \ref{sec:discrete_optimization}}

\begin{myproof}[of Lemma \ref{lem:geometric_gradient_decay}]
We proceed by induction on $m$ to prove part (a). The base case $m=0$ holds by assumption. Assume $x_m \in \mathcal{R}_{\epsilon}(f)$. 

Because $\epsilon \le \kappa$, we have $x_m \in \mathcal{R}_{\kappa}(f)$. By Property 2 of Definition \ref{def:function_class}, we have
\begin{equation}
\label{jacobian bound}
\sup_{u \in E_\perp(x_m), \|u\|=1} u^\top J_\xi(x_m) u \le -\gamma < 0.
\end{equation}

By definition, we have $\|x_{m+1} - x_m\|=\|\alpha \xi(x_m)\| \le \alpha \epsilon < d_{\partial}$, which implies that the line segment connecting $x_m$ and $x_{m+1}$ is contained in $\mathrm{int}(\mathcal{D})$. This validates the application of a Taylor expansion to $\xi(x_{m+1}) = \xi(x_m + \alpha \xi(x_m))$ around $x_m$, from which we have
\begin{equation}\label{eq:taylor_expansion}
\|\xi(x_{m+1})\| \le \|[I + \alpha J_\xi(x_m)]\xi(x_m)\| + M \alpha^2 \|\xi(x_m)\|^2.
\end{equation}
Since $\xi(x_m) = \Pi(x_m)\nabla f(x_m) \in E_\perp(x_m)$, \eqref{jacobian bound} implies
\begin{equation}\label{eq:linear_component_norm}
\begin{aligned} 
\|[I + \alpha J_\xi(x_m)]\xi(x_m)\|^2 &= \|\xi(x_m)\|^2 + 2\alpha \xi(x_m)^\top J_\xi(x_m) \xi(x_m) + \alpha^2 \|J_\xi(x_m) \xi(x_m)\|^2 \\ &\le (1 - 2\alpha \gamma + \alpha^2 L^2) \|\xi(x_m)\|^2 \\
& < (1 - \alpha\gamma)\|\xi(x_m)\|^2,
\end{aligned}
\end{equation}
where the last inequality follows from $\alpha < \frac{\gamma}{L^2}$. Because $(1 - \alpha\gamma) < (1 - \frac{\alpha\gamma}{2})^2$, we have
\begin{equation}\label{eq:linear_component_sqrt}
\|[I + \alpha J_\xi(x_m)]\xi(x_m)\| \le \left(1 - \frac{\alpha\gamma}{2}\right) \|\xi(x_m)\|.
\end{equation}
Substituting \eqref{eq:linear_component_sqrt} into \eqref{eq:taylor_expansion} yields
\begin{equation}\label{eq:recursive_bound}
\|\xi(x_{m+1})\| \le \left(1 - \frac{\alpha\gamma}{2} + M\alpha^2\|\xi(x_m)\|\right) \|\xi(x_m)\| \le \left(1 - \frac{\alpha\gamma}{2} + M\alpha^2\epsilon\right) \|\xi(x_m)\|,
\end{equation}
where we use the inductive hypothesis $\|\xi(x_m)\| \le \epsilon$. Because $\alpha < \frac{\gamma}{4 M \epsilon}$, it follows that $M\alpha^2\epsilon < \frac{\alpha\gamma}{4}$. Substituting this bound into \eqref{eq:recursive_bound} yields
\begin{equation}\label{eq:inductive_closure}
\|\xi(x_{m+1})\| < \left(1 - \frac{\alpha\gamma}{2} + \frac{\alpha\gamma}{4}\right) \|\xi(x_m)\| = \left(1 - \frac{\alpha\gamma}{4}\right) \|\xi(x_m)\| = \rho \|\xi(x_m)\|.
\end{equation}
Since $\rho < 1$, we have $\|\xi(x_{m+1})\| \le \rho \epsilon < \epsilon \le \kappa$. By Property 3 of Definition \ref{def:function_class}, $\mathrm{Re}(\mu_{k+1}(x_{m+1})) \le -\gamma/2 < 0$. Thus, $x_{m+1} \in \mathcal{R}_{\epsilon}(f)$, completing the induction for part (a). Notice that \eqref{eq:inductive_closure} establishes part (b).

For part (c), by definition and \eqref{eq:inductive_closure}, we have $\|x_{m+1} - x_m\| = \| \alpha \xi(x_m) \| \le \alpha \rho^m \|\xi(x_0)\|$, which leads to
\begin{equation}
\mathcal{L}_\alpha(x_0) = \sum_{m=0}^\infty \|x_{m+1} - x_m\| \le \alpha \|\xi(x_0)\| \sum_{m=0}^\infty \rho^m = \frac{\alpha}{1 - \rho} \|\xi(x_0)\|.
\end{equation}
Substituting $\rho = 1 - \frac{\alpha\gamma}{4}$ gives
\begin{equation}
\mathcal{L}_\alpha(x_0) \le \frac{\alpha}{1 - \left(1 - \frac{\alpha\gamma}{4}\right)} \|\xi(x_0)\| = \frac{4}{\gamma} \|\xi(x_0)\|.
\end{equation}
This establishes part (c) and concludes the proof.
\end{myproof}

\begin{myproof}[of Theorem \ref{thm:unified_convergence}]
By Lemma \ref{lem:geometric_gradient_decay}, the total discrete arc length $\mathcal{L}_\alpha(x_0) = \sum_{m=0}^\infty \|x_{m+1} - x_m\|$ is finite, which implies $\{x_m\}$ forms a Cauchy sequence in $\mathbb{R}^d$. Thus, the sequence $\{x_m\}$ converges to a unique limit point $x_\infty$. Because $x_m \in \mathcal{R}_{\epsilon}(f)$ for all $m$, and $\mathcal{R}_{\epsilon}(f)$ is a closed set, it must contain its limit points, yielding $x_\infty\in \mathcal{R}_{\epsilon}(f)$. By the continuity of $\xi$ and Lemma \ref{lem:geometric_gradient_decay}, we have $\|\xi(x_\infty)\| = \lim_{m \to \infty} \|\xi(x_m)\| = 0$.  Property 3 of Definition \ref{def:function_class} guarantees that $\mathrm{Re}(\mu_{k+1}(x_\infty)) \le -\gamma/2 < 0$, which establishes $x_\infty \in \mathcal{R}(f)$.

Using the triangle inequality and Lemma \ref{lem:geometric_gradient_decay}, we can write
\begin{equation}\label{eq:triangle_expansion}
\|x_m - x_\infty\| \le \sum_{j=m}^\infty \|x_{j+1} - x_j\| = \sum_{j=m}^\infty \alpha \|\xi(x_j)\| \le \alpha \|\xi(x_0)\| \sum_{j=m}^\infty \rho^j = \alpha \|\xi(x_0)\| \frac{\rho^m}{1 - \rho} .
\end{equation}
By substituting $\|\xi(x_0)\| \le \epsilon$ and utilizing the relation $1 - \rho = \frac{\alpha \gamma}{4}$, we obtain
\begin{equation}
\|x_m - x_\infty\| \le \alpha \|\xi(x_0)\| \frac{\rho^m}{\alpha \gamma / 4} = \frac{4 \|\xi(x_0)\|}{\gamma} \rho^m \le C \rho^m.
\end{equation}
\end{myproof}

\begin{myproof}[of Proposition \ref{prop:continuous_limit}]
Given $f \in C^4$, the vector field $\xi(x) = \Pi(x) \nabla f(x)$ is $C^2$-smooth, ensuring the one-step iteration map $G_\alpha(x)$ is continuous. By Lemma \ref{lem:geometric_gradient_decay}, $G_\alpha(x) \in \mathcal{R}_{\epsilon}(f)$ for all $x \in \mathcal{R}_{\epsilon}(f)$. Thus, the $m$-fold composition $G_\alpha^m(x)$ is a continuous function on $\mathcal{R}_{\epsilon}(f)$ for all $m \ge 0$.

By Theorem \ref{thm:unified_convergence}, for all $x \in \mathcal{R}_{\epsilon}(f)$ and $m \ge 0$,
$$ \|G_\alpha^m(x) - \Phi_\alpha(x)\| \le C \rho^m. $$
Since the bound $C \rho^m$ is independent of $x$ and $\lim_{m \to \infty} C \rho^m = 0$, the sequence $G_\alpha^m(x)$ converges uniformly to $\Phi_\alpha(x)$ on $\mathcal{R}_{\epsilon}(f)$. By the Uniform Limit Theorem, the uniform limit of continuous functions is continuous. Thus, $\Phi_\alpha(x)$ is continuous on $\mathcal{R}_{\epsilon}(f)$.
\end{myproof}

\begin{myproof}[of Proposition \ref{prop:discrete_limit_error}]
Recall $x_m = G_\alpha^m(x_0)$. By Theorem \ref{thm:unified_convergence}, $\lim_{m\to\infty}x_m = \Phi_\alpha(x_0) \in \mathcal{R}(f)$. Because $\Phi$ is continuous (see Lemma \ref{lem:lipschitz_projection}) and acts as the identity operator on $\mathcal{R}(f)$, we have $\lim_{m\to\infty} \Phi(x_m) = \Phi(\Phi_\alpha(x_0)) = \Phi_\alpha(x_0)$. Thus, we can write
\begin{equation}\label{eq:telescoping_sum}
    \|\Phi_\alpha(x_0) - \Phi(x_0)\| = \Big\| \sum_{m=0}^\infty \left[ \Phi(x_{m+1}) - \Phi(x_m) \right] \Big\| \le \sum_{m=0}^\infty \|\Phi(x_{m+1}) - \Phi(x_m)\|.
\end{equation}

For the continuous flow $\varphi_t$ defined in \eqref{varphi_def}, we have $\Phi(x_m) = \Phi(\varphi_\alpha(x_m))$. By definition, $x_{m+1} = x_m + \alpha \xi(x_m)$, which yields $\|x_{m+1} - x_m\| = \alpha \|\xi(x_m)\| \le \alpha \epsilon$. By Lemma \ref{lem:continuous_limit}, $\|\xi(\varphi_t(x_m))\| \le \|\xi(x_m)\| \le \epsilon$ for all $t \ge 0$, which leads to $\|\varphi_\alpha(x_m) - x_m\| \le \int_0^\alpha \|\xi(\varphi_t(x_m))\| \, dt \le \alpha \epsilon$. Thus, both $x_{m+1}$ and $\varphi_\alpha(x_m)$ belong to the closed Euclidean ball $B_{\alpha \epsilon}(x_m)$. 

For $y \in B_{\alpha \epsilon}(x_m)$, define the line segment $z(s) = x_m + s(y - x_m)$ for $s \in [0,1]$. Assume for contradiction that $z(s) \notin \mathcal{R}_{\kappa}(f)$ for some $s$. Since $z(0) = x_m \in \mathcal{R}_\epsilon(f)$ and $z(s)$ is continuous, there exists a first exit time $s_* \le 1$ such that $\|\xi(z(s_*))\| = \kappa$. The segment $z([0, s_*])$ is contained in $\mathcal{R}_{\kappa}(f) \subset \mathrm{int}(\mathcal{D})$. Applying the Mean Value Theorem along this segment yields
\begin{equation}
    \|\xi(z(s_*)) - \xi(x_m)\| \le L \|z(s_*) - x_m\| \le L \|y - x_m\| \le L \alpha \epsilon.
\end{equation}
By the triangle inequality, $\|\xi(z(s_*))\| \le \|\xi(x_m)\| + L \alpha \epsilon \le \epsilon + L \alpha \epsilon$. Because $\alpha < \frac{\kappa - \epsilon}{L \epsilon}$, we obtain $\|\xi(z(s_*))\| < \kappa$, which contradicts $\|\xi(z(s_*))\| = \kappa$. Thus, no such $s_*$ exists, yielding $B_{\alpha \epsilon}(x_m) \subset \mathcal{R}_{\kappa}(f)$.

Because $B_{\alpha \epsilon}(x_m)$ is convex, the line segment connecting $x_{m+1}$ and $\varphi_\alpha(x_m)$ is contained in $B_{\alpha \epsilon}(x_m) \subset \mathcal{R}_{\kappa}(f)$. By Lemma \ref{lem:lipschitz_projection}, $\|J_\Phi(x)\| \le L_\Phi$ for all $x \in \mathcal{R}_{\kappa}(f)$. Applying the Mean Value Theorem along this segment, we have
\begin{equation}\label{eq:lipschitz_bound}
    \|\Phi(x_{m+1}) - \Phi(x_m)\| = \|\Phi(x_{m+1}) - \Phi(\varphi_\alpha(x_m))\| \le L_\Phi \|x_{m+1} - \varphi_\alpha(x_m)\|.
\end{equation}

Since $\dot{\varphi}_t = \xi(\varphi_t)$, $\ddot{\varphi}_t = J_\xi(\varphi_t)\xi(\varphi_t)$ and $\|\xi(\varphi_t(x_m))\| \le \|\xi(x_m)\|$ by Lemma \ref{lem:continuous_limit}, we obtain
\begin{align}\label{eq:truncation_error}
     \|x_{m+1} - \varphi_\alpha(x_m)\| & = \|\varphi_\alpha(x_m) - (x_m + \alpha \xi(x_m))\| \nonumber\\
     & \le \frac{1}{2} \alpha^2 \sup_{t \in [0, \alpha]} \|\ddot{\varphi}_t\| \le \frac{1}{2} \alpha^2 L \sup_{t \in [0, \alpha]} \|\xi(\varphi_t(x_m))\| \le \frac{1}{2} L \alpha^2 \|\xi(x_m)\|.
\end{align}
Substituting \eqref{eq:truncation_error} into \eqref{eq:lipschitz_bound} and applying the result to \eqref{eq:telescoping_sum} yields
\begin{equation}\label{eq:substituted_sum}
    \|\Phi_\alpha(x_0) - \Phi(x_0)\| \le \sum_{m=0}^\infty \frac{1}{2} L_\Phi L \alpha^2 \|\xi(x_m)\| = \frac{1}{2} L_\Phi L \alpha \sum_{m=0}^\infty \alpha \|\xi(x_m)\| = \frac{1}{2} L_\Phi L \alpha \mathcal{L}_\alpha(x_0).
\end{equation}
%where $\mathcal{L}_\alpha(x_0) = \sum_{m=0}^\infty \alpha \|\xi(x_m)\|$. 
By Lemma \ref{lem:geometric_gradient_decay}, it follows that
\begin{equation}
    \|\Phi_\alpha(x_0) - \Phi(x_0)\| \le \frac{1}{2} L_\Phi L \alpha \left( \frac{4}{\gamma} \|\xi(x_0)\| \right) = \left( \frac{2 L_\Phi L}{\gamma} \|\xi(x_0)\| \right) \alpha,
\end{equation}
which concludes the proof.
\end{myproof}

\begin{myproof}[of Theorem \ref{thm:discrete_surjectivity}]
For $G_\alpha$ defined in \eqref{eq:discrete_operator}, its Jacobian is $J_{G_\alpha}(x) = I + \alpha J_\xi(x)$, $x \in \mathcal{D}$. Because $\alpha < 1/L$, we have $\|\alpha J_\xi(x)\| \le \alpha L < 1$ for all $x \in \mathcal{D}$. Thus, $J_{G_\alpha}(x)$ is non-singular on $\mathcal{D}$. By the Inverse Function Theorem, $G_\alpha$ is a $C^1$-smooth local diffeomorphism on $\mathcal{R}_{\epsilon}(f)$.

Let $x_* \in \mathcal{R}(f)$. Because $\xi(x_*) = 0$, we have $G_\alpha(x_*) = x_*$. The Jacobian of $G_\alpha$ at $x_*$ is $J_{G_\alpha}(x_*) = I + \alpha J_\xi(x_*)$. Since $J_\xi(x_*)$ has $k$ zero eigenvalues corresponding to the tangent space, $J_{G_\alpha}(x_*)$ has $k$ eigenvalues equal to $1$. 

Let $\mu$ be one of the $d-k$ non-zero eigenvalues of $J_\xi(x_*)$. The corresponding eigenvalue of $J_{G_\alpha}(x_*)$ is $1 + \alpha\mu$. Because $\mathrm{Re}(\mu) \le -\gamma$ and $|\mu| \le L$, the squared modulus satisfies 
\begin{equation}
    |1 + \alpha\mu|^2 \le 1 - 2\alpha\gamma + \alpha^2 L^2 < 1 - \alpha\gamma < \left(1 - \frac{\alpha\gamma}{4}\right)^2,
\end{equation}
where we use $\alpha^2 L^2 < \alpha\gamma$ due to $\alpha < \frac{\gamma}{L^2}$. In other words, $J_{G_\alpha}(x_*)$ has $k$ eigenvalues equal to $1$ and the remaining $d-k$ eigenvalues have moduli bounded from above by $\rho = 1 - \frac{\alpha\gamma}{4} < 1$. Then the Center Manifold Theorem for discrete-time systems \citep[Section 5.1.2]{kuznetsov2023elements} guarantees the existence of a $C^1$-smooth, $(d-k)$-dimensional manifold $W^s_{G_\alpha}(x_*)$ passing through $x_*$. For a local neighborhood $U$ of $x_*$, this manifold is defined as
\begin{equation}
    W^s_{G_\alpha}(x_*) = \Big\{ y \in U :\; \lim_{m \to \infty} G_\alpha^m(y) = x_* \Big\}.
\end{equation}
By definition of $\Phi_\alpha$, we have $\Phi_\alpha(y) = x_*$ for all $y \in W^s_{G_\alpha}(x_*)$.

Let $g(x) = \|\xi(x)\|$. Because $k < d$, we can choose $y \in W^s_{G_\alpha}(x_*) \setminus \{x_*\}$ such that $0 < g(y) < \epsilon$. Since $G_\alpha$ is a local diffeomorphism on $\mathcal{R}_{\epsilon}(f)$, its local inverse $G_\alpha^{-1}$ exists. Consider the sequence of backward iterates defined by $y_{-(m+1)} = G_\alpha^{-1}(y_{-m})$ with $y_0 = y$. 

Following the derivation in Lemma \ref{lem:geometric_gradient_decay}, while $y_{-(m+1)} \in \mathcal{R}_{\epsilon}(f)$, we have
\begin{equation}
    g(y_{-m}) = g(G_\alpha(y_{-(m+1)})) \le \rho g(y_{-(m+1)}),
\end{equation}
or equivalently, $g(y_{-(m+1)}) \ge \rho^{-1} g(y_{-m})$. By induction, we obtain $g(y_{-m}) \ge \rho^{-m} g(y)$. Because $\rho < 1$, $g(y_{-m})$ grows exponentially. Thus, the sequence $\{y_{-m}\}$ cannot remain in the sub-level set $\{x \in \mathcal{D} : g(x) \le \epsilon\}$ for all $m \ge 0$. There exists a finite integer $M > 0$ such that $x_{\text{out}} = y_{-M} \in W^s_{G_\alpha}(x_*)$ satisfies $g(x_{\text{out}}) > \epsilon$.

Since $W^s_{G_\alpha}(x_*)$ is diffeomorphic to an open ball in $\mathbb{R}^{d-k}$, it is path-connected. Let $\sigma: [0,1] \to W^s_{G_\alpha}(x_*)$ be a continuous path satisfying $\sigma(0) = x_*$ and $\sigma(1) = x_{\text{out}}$. Because $g(\sigma(s))$ is continuous with $g(\sigma(0)) = 0$ and $g(\sigma(1)) > \epsilon$, the Intermediate Value Theorem guarantees the existence of $s_0 \in (0,1)$ such that $x_0 = \sigma(s_0) \in W^s_{G_\alpha}(x_*)$ satisfies $g(x_0) = \epsilon$. 

Because $g(x_0) = \epsilon \le \kappa$, Property 3 of Definition \ref{def:function_class} implies $\mathrm{Re}(\mu_{k+1}(x_0)) \le -\gamma/2 < 0$, yielding $x_0 \in \partial \mathcal{R}_{\epsilon}(f)$. Because $x_0 \in W^s_{G_\alpha}(x_*)$, it follows that $\Phi_\alpha(x_0) = x_*$. Because $x_* \in \mathcal{R}(f)$ is arbitrary, $\Phi_\alpha$ is surjective.
\end{myproof}

\subsection{Proofs for Section \ref{sec:finite_sample_consistency}}

\begin{myproof}[of Lemma \ref{lem:perturbation_bounds}]
Assume the event $\mathcal{E}_n(c)$ holds. Define $\delta = \eta / 4$. Since the density derivative convergence rates satisfy $r_1(n) = o(r_2(n))$ and $r_2(n) = o(r_3(n))$, there exists an integer $N_0$ such that for all $n \ge N_0$, we have $r_1(n) \le r_2(n) \le r_3(n)$ and $d C_2 r_2(n) \le \delta$.

Part (a): By Lemma \ref{lem:kde_uniform_convergence_complete}, the event $\mathcal{E}_n(c)$ guarantees
$$ \|\nabla^2 \widehat{f}(x) - \nabla^2 f(x)\| \le d \max_{|a|=2} |D^a \widehat{f}(x) - D^a f(x)| \le d C_2 r_2(n),$$
which further implies via Weyl's inequality that
$$ \max_j |\widehat{\lambda}_j(x) - \lambda_j(x)| \le \|\nabla^2 \widehat{f}(x) - \nabla^2 f(x)\| \le d C_2 r_2(n). $$

Part (b): Because $\sup_{x \in \mathcal{D}}[\lambda_k(x) - \lambda_{k+1}(x)] \ge \eta > 0$, the Davis-Kahan $\sin \Theta$ theorem yields
$$ \|\widehat{\Pi}(x) - \Pi(x)\| \le \frac{2}{\eta} \|\nabla^2 \widehat{f}(x) - \nabla^2 f(x)\| \le \frac{2 d C_2}{\eta} r_2(n), \quad \text{for all } x \in \mathcal{D}. $$

Part (c): Using the definition of $\xi$ and $\widehat{\xi}$, it follows from the triangle inequality that
$$ \|\widehat{\xi}(x) - \xi(x)\| \le \|\widehat{\Pi}(x)(\nabla \widehat{f}(x) - \nabla f(x))\| + \|(\widehat{\Pi}(x) - \Pi(x))\nabla f(x)\|. $$
Because $\|\widehat{\Pi}(x)\| = 1$ and Assumption A1 guarantees the existence of a constant $b_1>0$ such that $\|\nabla f(x)\| \le b_1$ for all $x\in\mathcal{D}$, we obtain
$$ \|\widehat{\xi}(x) - \xi(x)\| \le \sqrt{d} C_1 r_1(n) + \left( \frac{2 d C_2}{\eta} r_2(n) \right) b_1. $$
We define the constant $C_\xi = \sqrt{d} C_1 + \frac{2 d C_2 b_1}{\eta}$. Because $n \ge N_0$ ensures $r_1(n) \le r_2(n)$,  we have $\|\widehat{\xi}(x) - \xi(x)\| \le C_\xi r_2(n)$.

Part (d): The Jacobian of $\xi(x) = \Pi(x)\nabla f(x)$ is given by
$$ J_\xi(x) = \Pi(x) \nabla^2 f(x) + \mathcal{T}(\nabla \Pi(x), \nabla f(x)), $$
where the tensor operator $\mathcal{T}$ produces a $d \times d$ matrix with entries $[\mathcal{T}(\nabla \Pi, \nabla f)]_{i,k} = \sum_{j=1}^d \frac{\partial \Pi_{i,j}}{\partial x_k} \frac{\partial f}{\partial x_j}$. 

Applying the triangle inequality, we get
\begin{align}\label{eq:jacobian_perturbation_triangle}
    \|J_{\widehat{\xi}}(x) - J_\xi(x)\| &\le \|\widehat{\Pi} \nabla^2 \widehat{f} - \Pi \nabla^2 f\| + \|\mathcal{T}(\nabla \widehat{\Pi}, \nabla \widehat{f}) - \mathcal{T}(\nabla \Pi, \nabla f)\| \nonumber\\
    &\le \|\widehat{\Pi} \nabla^2 \widehat{f} - \Pi \nabla^2 f\| + \|\mathcal{T}(\nabla \widehat{\Pi} - \nabla \Pi, \nabla \widehat{f})\| + \|\mathcal{T}(\nabla \Pi, \nabla \widehat{f} - \nabla f)\|.
\end{align}
Because $\mathcal{D}$ is compact and $f \in C^4$, $\|\nabla \Pi\|$ and $\|\nabla f\|$ are uniformly bounded on $\mathcal{D}$. Because $\nabla \widehat{f}$ converges uniformly to $\nabla f$ by Lemma \ref{lem:kde_uniform_convergence_complete}, it is uniformly bounded for all $n \ge N_0$. 

Let $\mathcal{S}_d$ denote the space of $d \times d$ symmetric matrices, and let $\mathcal{P}: \mathcal{S}_d \to \mathbb{R}^{d \times d}$ denote the projection mapping onto the subspace spanned by the top $k$ eigenvectors, such that $\Pi(x) = \mathcal{P}(\nabla^2 f(x))$ and $\widehat{\Pi}(x) = \mathcal{P}(\nabla^2 \widehat{f}(x))$. Because $\lambda_k - \lambda_{k+1}$ is uniformly bounded below by $\eta$, the image set $\nabla^2 f(\mathcal{D})$ is a compact subset of $\mathcal{S}_d$. Define the compact $\delta$-tube $\mathcal{U}_\delta = \{ A \in \mathcal{S}_d : \inf_{x \in \mathcal{D}} \|A - \nabla^2 f(x)\| \le \delta \}$. Because $\delta = \eta / 4$, Weyl's inequality guarantees every matrix in $\mathcal{U}_\delta$ has a gap between the $k$th and $(k+1)$th eigenvalues strictly bounded below by $\eta/2 > 0$. On this domain, the first and second differentials of $\mathcal{P}$, denoted $D\mathcal{P}$ and $D^2\mathcal{P}$, exist and are uniformly bounded in operator norm by finite constants $c_\eta$ and $L_\eta$ on $\mathcal{U}_\delta$.

Applying the chain rule, the partial derivative of $\Pi$ with respect to the $i$-th coordinate $x_i$ is $\partial_{x_i} \Pi(x) = D\mathcal{P}(\nabla^2 f(x))[\partial_{x_i} \nabla^2 f(x)]$. Using the triangle inequality, we have
\begin{align}\label{eq:differential_triangle}
    \|\partial_{x_i} \widehat{\Pi}(x) - \partial_{x_i} \Pi(x)\| &\le \|D\mathcal{P}(\nabla^2 \widehat{f}(x)) - D\mathcal{P}(\nabla^2 f(x))\| \|\partial_{x_i} \nabla^2 \widehat{f}(x)\| \nonumber \\
    &\quad + \|D\mathcal{P}(\nabla^2 f(x))\| \|\partial_{x_i} \nabla^2 \widehat{f}(x) - \partial_{x_i} \nabla^2 f(x)\|.
\end{align}

Since $n \ge N_0$ is large enough that $\|\nabla^2 \widehat{f}(x) - \nabla^2 f(x)\| \le \delta$ for all $x \in \mathcal{D}$, the straight line segment connecting $\nabla^2 f(x)$ and $\nabla^2 \widehat{f}(x)$ lies within a closed, convex $\delta$-ball in $\mathcal{S}_d$ centered at $\nabla^2 f(x)$, which is contained in $\mathcal{U}_\delta$. Applying the Mean Value Theorem along this convex segment yields
$$ \|D\mathcal{P}(\nabla^2 \widehat{f}(x)) - D\mathcal{P}(\nabla^2 f(x))\| \le \sup_{A \in \mathcal{U}_\delta} \|D^2\mathcal{P}(A)\| \|\nabla^2 \widehat{f}(x) - \nabla^2 f(x)\| \le L_\eta (d C_2 r_2(n)). $$
Bounding the remaining terms in \eqref{eq:differential_triangle} we obtain
$$ \|\partial_{x_i} \widehat{\Pi}(x) - \partial_{x_i} \Pi(x)\| \le L_\eta (d C_2 r_2(n)) \left(\|\partial_{x_i} \nabla^2 f(x)\| + d^2 C_3 r_3(n)\right) + c_\eta (d^2 C_3 r_3(n)). $$
Given $f \in C^4$, $\|\partial_{x_i} \nabla^2 f(x)\|$ is uniformly bounded on $\mathcal{D}$. Therefore, there exists a constant $C_{\Pi'} > 0$ such that $\sup_{x \in \mathcal{D}} \|\nabla \widehat{\Pi}(x) - \nabla \Pi(x)\| \le C_{\Pi'} r_3(n)$ by recalling that $n \ge N_0$ ensures $r_2(n) \le r_3(n)$. Substituting these bounds into \eqref{eq:jacobian_perturbation_triangle} establishes the existence of a constant $C_J > 0$ satisfying
$$ \sup_{x \in \mathcal{D}} \|J_{\widehat{\xi}}(x) - J_\xi(x)\| \le C_J r_3(n). $$
\end{myproof}

\begin{myproof}[Proofs of Corollary \ref{cor:empirical_assumptions}]
Let event $\mathcal{E}_n(c)$ hold. Recall $L = \sup_{z \in \mathcal{D}} \|J_\xi(z)\|$. Define $\epsilon_\Pi(n) = \frac{2 d C_2}{\eta} r_2(n)$, $\epsilon_Q(n) = 2\gamma \epsilon_\Pi(n) + 2 L \epsilon_\Pi(n) + L \epsilon_\Pi(n)^2$, and $C_\mu = (2L + 1)^{1 - 1/d}$. Because $r_\ell(n) \to 0$ for $\ell=0,\dots,4$, and $\epsilon_\xi(n), \epsilon_J(n), \epsilon_Q(n) \to 0$, we can choose $N_1 \ge N_0$ such that for all $n \ge N_1$,
$$ \epsilon_J(n) \le \min\left(1, \left(\frac{\gamma}{5 C_\mu}\right)^d\right), \quad \epsilon_\xi(n) < \frac{\kappa}{2}, \quad 2 d C_2 r_2(n) \le \frac{\eta}{2}, $$
$$ \epsilon_Q(n) + \epsilon_J(n) \le \frac{\gamma}{2}, \quad \max_{0 \le \ell \le 4} C_\ell r_\ell(n) \le 1. $$

Part (a): By Assumption (B1), we have $K \in C^4(\mathbb{R}^d, \mathbb{R})$, which implies $\widehat{f} \in C^4(\mathbb{R}^d, \mathbb{R})$. Under Assumption  (A1), $\max_{|a|=\ell} \sup_{x \in \mathcal{D}} |D^a f(x)| \le b_\ell$ for some constants $b_\ell>0$, $\ell=0,1,2,3,4$. By Lemma \ref{lem:kde_uniform_convergence_complete} and the triangle inequality, we have $\max_{|a|=\ell} \sup_{x \in \mathcal{D}} |D^a \widehat{f}(x)| \le b_\ell + C_\ell r_\ell(n)\le b_\ell+1 =:B_\ell,$ because $n \ge N_1$ ensures $C_\ell r_\ell(n) \le 1$.

Part (b): By Lemma \ref{lem:perturbation_bounds}, part (a), for all $x \in \mathcal{D}$,
$$ \widehat{\lambda}_k(x) - \widehat{\lambda}_{k+1}(x) \ge \eta - 2 d C_2 r_2(n). $$
For $n \ge N_1$, $2 d C_2 r_2(n) \le \eta / 2$, leading to $\widehat{\lambda}_k(x) - \widehat{\lambda}_{k+1}(x) \ge \eta / 2$.

Part (c): Let $x \in \mathcal{R}_{\kappa/2}(\widehat{f})$. By definition, $\|\widehat{\xi}(x)\| \le \kappa/2$. By the triangle inequality and Lemma \ref{lem:perturbation_bounds}, part (c),
$$ \|\xi(x)\| \le \|\widehat{\xi}(x)\| + \epsilon_\xi(n) < \frac{\kappa}{2} + \frac{\kappa}{2} = \kappa,$$
which implies $\mathrm{Re}(\mu_{k+1}(x)) \le -\gamma/2$, by Property 3 of Definition \ref{def:function_class}. Because $\|\xi(x)\| \le \kappa$ and $\mathrm{Re}(\mu_{k+1}(x)) \le -\gamma/2$, we obtain $x \in \mathcal{R}_{\kappa}(f)$.

Let $u \in \widehat{E}_{\perp}(x)$ with $\|u\| = 1$, which guarantees $\widehat{\Pi}(x) u = u$. Decompose $u = v + w$ for $v = \Pi(x)u \in E_\perp(x)$ and $w = (I - \Pi(x))u \in E_\parallel(x)$, such that $1=\|u\|\le \|v\| + \|w\|$. By Lemma \ref{lem:perturbation_bounds}, part (b), 
$$ \|w\| = \|u - \Pi(x) u\| = \|\widehat{\Pi}(x) u - \Pi(x) u\| \le \|\widehat{\Pi}(x) - \Pi(x)\| \|u\| \le \epsilon_\Pi(n). $$
Thus, $1 - \epsilon_\Pi(n) \le 1- \|w\| \le \|v\| \le 1$. It is clear that
\begin{align}
\label{uJu expansion}
 u^\top J_\xi(x) u = v^\top J_\xi(x) v + 2 v^\top J_\xi(x) w + w^\top J_\xi(x) w. 
 \end{align}
Because $v \in E_\perp(x)$ and $x \in \mathcal{R}_{\kappa}(f)$, it follows from Property 2 of Definition \ref{def:function_class} that $$v^\top J_\xi(x) v \le -\gamma \|v\|^2 \le -\gamma (1 - \epsilon_\Pi(n))^2 \le -\gamma + 2\gamma \epsilon_\Pi(n).$$ 
By the Cauchy-Schwarz inequality, 
$$ 2|v^\top J_\xi(x) w| \le 2 \|v\| \|J_\xi(x)\| \|w\| \le 2 L \epsilon_\Pi(n), $$
$$ |w^\top J_\xi(x) w| \le \|w\| \|J_\xi(x)\| \|w\| \le L \epsilon_\Pi(n)^2. $$
Substituting these components into \eqref{uJu expansion} yields
$$ u^\top J_\xi(x) u \le -\gamma + 2\gamma \epsilon_\Pi(n) + 2 L \epsilon_\Pi(n) + L \epsilon_\Pi(n)^2 = -\gamma + \epsilon_Q(n). $$
Then by Lemma \ref{lem:perturbation_bounds}, part (d),
$$ u^\top J_{\widehat{\xi}}(x) u = u^\top J_\xi(x) u + u^\top (J_{\widehat{\xi}}(x) - J_\xi(x)) u \le -\gamma + \epsilon_Q(n) + \epsilon_J(n). $$
Since $n \ge N_1$ is large enough that $\epsilon_Q(n) + \epsilon_J(n) \le \gamma/2$, taking the supremum over $u \in \widehat{E}_{\perp}(x)$ we get the desired result in part (c).

Part (d): Suppose $x \in \mathcal{D}$ satisfies $\|\widehat{\xi}(x)\| < \kappa/2$. By Lemma \ref{lem:perturbation_bounds}, part (c),
$$ \|\xi(x)\| \le \|\widehat{\xi}(x)\| + \|\widehat{\xi}(x) - \xi(x)\| < \frac{\kappa}{2} + \frac{\kappa}{2} = \kappa. $$
By Property 3 of Definition \ref{def:function_class}, $\|\xi(x)\| < \kappa$ implies $\mathrm{Re}(\mu_{k+1}(x)) \le -\gamma/2$. 

By definition, the sequences $\{\mu_j(x)\}_{j=1}^d$ and $\{\widehat{\mu}_j(x)\}_{j=1}^d$ are sorted in descending order of their real parts. By Elsner's Theorem \citep{elsner1985optimal}, there exists a bijection $\pi$ on the indices $\{1, \dots, d\}$ such that
$$ \max_j |\widehat{\mu}_{\pi(j)}(x) - \mu_j(x)| \le \big( \|J_{\widehat{\xi}}(x)\| + \|J_\xi(x)\| \big)^{1 - 1/d} \|J_{\widehat{\xi}}(x) - J_\xi(x)\|^{1/d}. $$
By Lemma \ref{lem:perturbation_bounds}, part (d), we have $\|J_{\widehat{\xi}}(x) - J_\xi(x)\| \le \epsilon_J(n)$. Because $n \ge N_1$ ensures $\epsilon_J(n) \le 1$, we have $\|J_{\widehat{\xi}}(x)\| \le L + 1$. Thus, 
$$ \max_j |\widehat{\mu}_{\pi(j)}(x) - \mu_j(x)| \le (2L + 1)^{1 - 1/d} \epsilon_J(n)^{1/d} = C_\mu \epsilon_J(n)^{1/d}  \le \gamma/5 < \gamma/4. $$
For $j \le k$, $\mathrm{Re}(\mu_j(x)) = 0$ implies $\mathrm{Re}(\widehat{\mu}_{\pi(j)}(x)) \ge -\gamma/5$. For $j > k$, $\mathrm{Re}(\mu_j(x)) \le -\gamma/2$ implies $\mathrm{Re}(\widehat{\mu}_{\pi(j)}(x)) \le -3\gamma/10$. Hence, $\{\pi(j) : j=1,\dots,k\} = \{1,\dots,k\}$ and $\{\pi(j) : j=k+1,\dots,d\} = \{k+1,\dots,d\}$. Consequently, there exists some $j' \ge k+1$ such that $|\widehat{\mu}_{k+1}(x) - \mu_{j'}(x)| \le \gamma/5$, and
$$ \mathrm{Re}(\widehat{\mu}_{k+1}(x)) \le \mathrm{Re}(\mu_{j'}(x)) + \frac{\gamma}{5} \le \mathrm{Re}(\mu_{k+1}(x)) + \frac{\gamma}{5} \le -\frac{\gamma}{2} + \frac{\gamma}{5} = -\frac{3\gamma}{10} < -\frac{\gamma}{4}. $$
We have shown that $\|\widehat{\xi}(x)\| < \kappa/2$ implies $\mathrm{Re}(\widehat{\mu}_{k+1}(x)) < -\gamma/4$ for $x\in\mathcal{D}$. Equivalently, this means $\mathrm{Re}(\widehat{\mu}_{k+1}(x)) \ge -\gamma/4$ implies $\|\widehat{\xi}(x)\| \ge \kappa/2$, concluding the proof of part (d).
\end{myproof}

\begin{myproof}[of Theorem \ref{thm:hausdorff_consistency}]
Let event $\mathcal{E}_n(c)$ hold. By Assumption (A1), $f \in \mathcal{F}(\eta, \gamma, \kappa, \mathcal{D})$. Recall $L = \sup_{z \in \mathcal{D}} \|J_\xi(z)\|$ and $M = \frac{1}{2} \sup_{z \in \mathcal{D}} \|\nabla^2 \xi(z)\|$. By Lemma \ref{lem:kde_uniform_convergence_complete}, we can choose $N_2 \ge N_1$ sufficiently large such that for all $n \ge N_2$, $\widehat{L} := \sup_{z \in \mathcal{D}} \|J_{\widehat{\xi}}(z)\|$ satisfies $L/2 \le \widehat{L} \le 2L$, and $\widehat{M} := \frac{1}{2} \sup_{z \in \mathcal{D}} \|\nabla^2 \widehat{\xi}(z)\| \le 2M$.

Furthermore, let $N_2$ be large enough that
\begin{equation}
\label{epsilon xi bound}
    \epsilon_\xi(n) \le \min\left( \frac{\kappa}{2}, \frac{L\gamma}{8M}, \frac{\gamma \kappa}{8L} \right).
\end{equation}
Because $n \ge N_2 \ge N_1$, Corollary \ref{cor:empirical_assumptions} establishes $\widehat{f} \in \mathcal{F}\left(\frac{\eta}{2}, \frac{\gamma}{2}, \frac{\kappa}{2}, \mathcal{D}\right)$. 

Let $\widehat{x} \in \mathcal{R}(\widehat{f})$. By definition, $\widehat{\xi}(\widehat{x}) = 0$, which implies $\|\xi(\widehat{x})\| = \|\xi(\widehat{x}) - \widehat{\xi}(\widehat{x})\| \le \epsilon_\xi(n)$. Because $\epsilon_\xi(n) \le \kappa/2 < \kappa$, Property 3 of Definition \ref{def:function_class} implies $\mathrm{Re}(\mu_{k+1}(\widehat{x})) \le -\gamma/2$. Thus, $\widehat{x} \in \mathcal{R}_{\epsilon_\xi(n)}(f)$, establishing $\mathcal{R}(\widehat{f}) \subseteq \mathcal{R}_{\epsilon_\xi(n)}(f)$. 
Using \eqref{epsilon xi bound}, we can apply Corollary \ref{cor:hausdorff_bound} to obtain
\begin{equation}
\label{eq:directed_hausdorff_1}
    d(\mathcal{R}(\widehat{f}) | \mathcal{R}(f)) \le d(\mathcal{R}_{\epsilon_\xi(n)}(f) | \mathcal{R}(f)) \le \frac{\epsilon_\xi(n)}{\gamma}.
\end{equation}

Let $x \in \mathcal{R}(f)$. By definition, we have $\xi(x) = 0$, and hence $\|\widehat{\xi}(x)\| \le \epsilon_\xi(n)$. Because $\epsilon_\xi(n) \le \kappa/2$, Corollary \ref{cor:empirical_assumptions} implies $\mathrm{Re}(\widehat{\mu}_{k+1}(x)) \le -\gamma/4$. Thus, $x \in \mathcal{R}_{\epsilon_\xi(n)}(\widehat{f})$, establishing $\mathcal{R}(f) \subseteq \mathcal{R}_{\epsilon_\xi(n)}(\widehat{f})$. 

By our choice of $N_2$, we have 
\begin{equation}
    \min\left( \frac{\kappa}{2}, \frac{\widehat{L}(\gamma/2)}{\widehat{M}}, \frac{(\gamma/2)}{\widehat{L}}\frac{\kappa}{2} \right) \ge \min\left( \frac{\kappa}{2}, \frac{(L/2)(\gamma/2)}{2M}, \frac{\gamma/2}{2L}\frac{\kappa}{2} \right) = \min\left( \frac{\kappa}{2}, \frac{L\gamma}{8M}, \frac{\gamma\kappa}{8L} \right) \ge \epsilon_\xi(n).
\end{equation}
This allows to apply Corollary \ref{cor:hausdorff_bound} to $\widehat{f}$, resulting in
\begin{equation}
\label{eq:directed_hausdorff_2}
    d(\mathcal{R}(f) | \mathcal{R}(\widehat{f})) \le d(\mathcal{R}_{\epsilon_\xi(n)}(\widehat{f}) | \mathcal{R}(\widehat{f})) \le \frac{\epsilon_\xi(n)}{\gamma/2} = \frac{2\epsilon_\xi(n)}{\gamma}.
\end{equation}

Combining the bounds from \eqref{eq:directed_hausdorff_1} and \eqref{eq:directed_hausdorff_2} yields
\begin{equation}
    d_H(\mathcal{R}(\widehat{f}), \mathcal{R}(f)) \le \max\left( \frac{\epsilon_\xi(n)}{\gamma}, \frac{2\epsilon_\xi(n)}{\gamma} \right) = \frac{2\epsilon_\xi(n)}{\gamma}.
\end{equation}
Substituting $\epsilon_\xi(n) = O(r_2(n))$ establishes the final asymptotic rate.
\end{myproof}

\begin{myproof}[of Theorem \ref{thm:empirical_scms_convergence}]
Assume the event $\mathcal{E}_n(c)$ holds for $n \ge N_2$. By Theorem \ref{thm:hausdorff_consistency}, $N_2$ ensures $\widehat{L} \le 2L$, $\widehat{M} \le 2M$, and $\epsilon_\xi(n) \le \kappa/2$.

Let $x \in \mathcal{R}_{\epsilon}(\widehat{f})$. By the triangle inequality and Lemma \ref{lem:perturbation_bounds},
\begin{equation}
    \|\xi(x)\| \le \|\widehat{\xi}(x)\| + \epsilon_\xi(n) \le \epsilon + \epsilon_\xi(n).
\end{equation}
Since $\epsilon \le \kappa/2$ and $\epsilon_\xi(n) \le \kappa/2$, we obtain $\|\xi(x)\| \le \kappa$. This establishes $\mathcal{R}_\epsilon(\widehat{f}) \subseteq \mathcal{R}_{\kappa}(f)$. Consequently, 
\begin{equation}
    \inf_{x \in \mathcal{R}_\epsilon(\widehat{f}), y \in \partial \mathcal{D}} \|x - y\| \ge \inf_{x \in \mathcal{R}_{\kappa}(f), y \in \partial \mathcal{D}} \|x - y\| = d_{\partial}.
\end{equation}
By Corollary \ref{cor:empirical_assumptions}, $\widehat{f}$ satisfies Assumption (A1) and $ \widehat{f} \in \mathcal{F}\left(\frac{\eta}{2}, \frac{\gamma}{2}, \frac{\kappa}{2}, \mathcal{D}\right).$ On the other hand, we can show the step size $\alpha$ satisfies the conditions in Theorem \ref{thm:unified_convergence} when $f$ is replaced by $\widehat{f}$, that is,
\begin{equation}
\label{alpha requirement}
    \alpha < \frac{\gamma}{8 L^2} \le \frac{\gamma/2}{2 \widehat{L}^2}, \quad \alpha < \frac{\gamma}{16 M \epsilon} \le \frac{\gamma/2}{4 \widehat{M} \epsilon}, \quad \text{and} \quad \alpha < \frac{d_{\partial}}{\epsilon}.
\end{equation}

Therefore, by invoking Theorem \ref{thm:unified_convergence}, we can show the convergence of $\{\widehat{x}_m\}_{m=0}^\infty$ to a unique limit point $\widehat{x}_\infty \in \mathcal{R}(\widehat{f})$ with the uniform bound
\begin{equation}
    \|\widehat{x}_m - \widehat{x}_\infty\| \le \widetilde{C} \widetilde{\rho}^m,
\end{equation}
where $\widetilde{C} = \frac{4\epsilon}{\gamma/2} = \frac{8\epsilon}{\gamma}$ and $\widetilde{\rho} = 1 - \frac{\alpha (\gamma/2)}{4} = 1 - \frac{\alpha \gamma}{8}$.
\end{myproof}

\begin{myproof}[of Theorem \ref{thm:empirical_discrete_surjectivity}]
Assume the event $\mathcal{E}_n(c)$ holds for $n \ge N_2$. By Theorem \ref{thm:hausdorff_consistency}, $N_2$ ensures $\widehat{L} \le 2L$, $\widehat{M} \le 2M$, and $\epsilon_\xi(n) \le \kappa/2$. By Corollary \ref{cor:empirical_assumptions}, we have $\widehat{f} \in \mathcal{F}\left(\frac{\eta}{2}, \frac{\gamma}{2}, \frac{\kappa}{2}, \mathcal{D}\right)$.

As established in Theorem \ref{thm:empirical_scms_convergence}, the step size $\alpha$ satisfies the bounds in \eqref{alpha requirement} for the empirical parameters $\widehat{L}$ and $\widehat{M}$. Furthermore, the step size satisfies $\alpha < \frac{1}{2L} \le \frac{1}{\widehat{L}}$, ensuring $\widehat{G}_\alpha(x) = x + \alpha \widehat{\xi}(x)$ is a local diffeomorphism on $\mathcal{D}$. 

Applying Theorem \ref{thm:discrete_surjectivity} to the vector field $\widehat{\xi}$, we conclude that that the limit map $\widehat{\Phi}_\alpha(x)$ is surjective from $\partial \mathcal{R}_{\epsilon}(\widehat{f})$ onto $\mathcal{R}(\widehat{f})$.
\end{myproof}

\begin{myproof}[of Corollary \ref{cor:algorithmic_hausdorff}]
Let $\widetilde{C} = \frac{8\epsilon}{\gamma}$ and $\widetilde{\rho} = 1 - \frac{\alpha \gamma}{8}$.

Let $\widehat{x}_0 \in \partial \mathcal{R}_{\epsilon}(\widehat{f})$. By Theorem \ref{thm:empirical_scms_convergence}, we have $\widehat{\Phi}_\alpha(\widehat{x}_0) \in \mathcal{R}(\widehat{f})$ and $\|\widehat{G}_\alpha^m(\widehat{x}_0) - \widehat{\Phi}_\alpha(\widehat{x}_0)\| \le \widetilde{C} \widetilde{\rho}^m$. Hence,
\begin{equation}\label{eq:hausdorff_forward}
    d(\widehat{\mathcal{X}}_m | \mathcal{R}(\widehat{f})) \le \sup_{\widehat{x}_0 \in \partial \mathcal{R}_{\epsilon}(\widehat{f})} \|\widehat{G}_\alpha^m(\widehat{x}_0) - \widehat{\Phi}_\alpha(\widehat{x}_0)\| \le \widetilde{C} \widetilde{\rho}^m.
\end{equation}

Following from Theorem \ref{thm:empirical_discrete_surjectivity}, $\widehat{\Phi}_\alpha$ is surjective from $\partial \mathcal{R}_{\epsilon}(\widehat{f})$ onto $\mathcal{R}(\widehat{f})$, that is, for every $x_* \in \mathcal{R}(\widehat{f})$, there exists $\widehat{x}_0 \equiv  \widehat{x}_0(x_*)\in \partial \mathcal{R}_{\epsilon}(\widehat{f})$ such that $\widehat{\Phi}_\alpha(\widehat{x}_0) = x_*$. By Theorem \ref{thm:empirical_scms_convergence}, $\|\widehat{G}_\alpha^m(\widehat{x}_0) - x_*\| \le \widetilde{C} \widetilde{\rho}^m$. Hence
\begin{equation}\label{eq:hausdorff_covering}
    d(\mathcal{R}(\widehat{f}) | \widehat{\mathcal{X}}_m) \le \sup_{x_* \in \mathcal{R}(\widehat{f})} \|\widehat{G}_\alpha^m(\widehat{x}_0) - x_*\| \le \widetilde{C} \widetilde{\rho}^m.
\end{equation}

Combining \eqref{eq:hausdorff_forward} and \eqref{eq:hausdorff_covering} establishes $d_H(\widehat{\mathcal{X}}_m, \mathcal{R}(\widehat{f})) \le \widetilde{C} \widetilde{\rho}^m$.
\end{myproof}

\begin{myproof}[of Theorem \ref{thm:total_error_bound}]
Let $n \ge N_2$. By Lemma \ref{lem:kde_uniform_convergence_complete}, the event $\mathcal{E}_n(c)$ occurs with probability at least $1 - n^{-c}$. Assume $\mathcal{E}_n(c)$ holds. By the triangle inequality for the Hausdorff metric,
\begin{equation}\label{eq:hausdorff_triangle}
    d_H(\widehat{\mathcal{X}}_m, \mathcal{R}(f)) \le d_H(\widehat{\mathcal{X}}_m, \mathcal{R}(\widehat{f})) + d_H(\mathcal{R}(\widehat{f}), \mathcal{R}(f)).
\end{equation}
By Corollary \ref{cor:algorithmic_hausdorff}, we have $d_H(\widehat{\mathcal{X}}_m, \mathcal{R}(\widehat{f})) \le \frac{8\epsilon}{\gamma} \left(1 - \frac{\alpha \gamma}{8}\right)^m$. By Theorem \ref{thm:hausdorff_consistency}, we have $d_H(\mathcal{R}(\widehat{f}), \mathcal{R}(f)) \le \frac{2 \epsilon_\xi(n)}{\gamma}$. Substituting these bounds into \eqref{eq:hausdorff_triangle} we obtain the final inequality.
\end{myproof}

\subsection{Proofs for Section \ref{sec:original_scms_log_density}}

\begin{myproof}[of Lemma \ref{lem:automatic_stability}]
Let $L_{\log} = \sup_{x \in \mathcal{D}} \|J_{\xi^{\log}}(x)\|$ and $M_{\log} = \frac{1}{2} \sup_{x \in \mathcal{D}} \|\nabla^2 \xi^{\log}(x)\|$. Under Assumptions (A1) and (A2$^\prime$), we have $p \in C^4$, $L_{\log}<\infty$ and $M_{\log}<\infty$. Also, similar to $d_{\partial,}$, we define $d_{\partial, \log} = \inf_{x \in \mathcal{R}_{\kappa^{\log}}(p), y \in \partial \mathcal{D}} \|x - y\| > 0$.
Let $N_3$ be sufficiently large such that for all $n \ge N_3$, conditionally on $\mathcal{E}_n(c)$, it satisfies that $\epsilon_\xi^{\log}(n) \le \kappa_{\log}/2$, $\Delta_n \le 1/2$, and
\begin{equation}
\label{varying step size}
    2 C_K h^2 < \min\left( \frac{\gamma_{\log}}{8 L_{\log}^2}, \; \frac{\gamma_{\log}}{8 M_{\log} \kappa_{\log}}, \; \frac{1}{2 L_{\log}}, \; \frac{2d_{\partial, \log}}{\kappa_{\log}} \right),
\end{equation}
and $\|\nabla \alpha_n(x)\| \le \frac{1}{2\kappa_{\log}}$ uniformly on $\mathcal{D}$. Assume the event $\mathcal{E}_n(c)$ holds for $n \ge N_3$.

The adaptive step size is $\alpha_n(x) = C_K h^2 \frac{\widehat{f}(x)}{\widehat{q}(x)}$. Since $\Delta_n = \sup_{x \in \mathcal{D}} \left|\frac{\widehat{f}(x)}{\widehat{q}(x)} - 1\right| \le 1/2$, the step size satisfies the uniform bounds
\begin{equation}\label{eq:adaptive_step_bounds}
    0 < \frac{C_K h^2}{2} \le C_K h^2 (1 - \Delta_n) \le \alpha_n(x) \le C_K h^2 (1 + \Delta_n) \le 2 C_K h^2.
\end{equation}
By \eqref{varying step size}, the upper bound $2 C_K h^2$ satisfies the step-size conditions of Theorem \ref{thm:empirical_discrete_surjectivity}. 

The Jacobian of $\widehat{G}^{\log} \equiv \widehat{G}_{\alpha_n}^{\log}$ at $x$ is
\begin{equation}
    J_{\widehat{G}^{\log}}(x) = I + \alpha_n(x) J_{\widehat{\xi}^{\log}}(x) + \widehat{\xi}^{\log}(x) \nabla \alpha_n(x)^{\top}.
\end{equation}

For any $x \in \mathcal{R}_\epsilon(\widehat{p})$, $\|\alpha_n(x) J_{\widehat{\xi}^{\log}}(x)\| \le 2 C_K h^2  L_{\log} \le 1/2$. 
In addition, since $\|\widehat{\xi}^{\log}(x)\| \le \epsilon \le\kappa_{\log}/2$, we have $\|\widehat{\xi}^{\log}(x) \nabla \alpha_n(x)^{\top}\| \le (\kappa_{\log}/2) \|\nabla \alpha_n(x)\|\le 1/4$, by the definition of $N_3$. These bounds ensure $J_{\widehat{G}^{\log}}(x)$ remains invertible on $\mathcal{R}_\epsilon(\widehat{p})$, establishing $\widehat{G}^{\log}$ as a local diffeomorphism. We are then able to apply analogous arguments for the proof Theorem \ref{thm:empirical_discrete_surjectivity} to $\widehat{G}^{\log}$ and prove that $\widehat{\Phi}^{\log}:=  \lim_{m \to \infty} (\widehat{G}^{\log})^m$ exists and is surjective from $\partial \mathcal{R}_{\epsilon}(\widehat{p})$ onto $\mathcal{R}(\widehat{p})$.
\end{myproof}

\begin{myproof}[of Theorem \ref{thm:scms_total_convergence}]
We use the same definition of $N_3$ and $\mathcal{E}_n(c)$ in the proof of Lemma \ref{lem:automatic_stability}, where the event $\mathcal{E}_n(c)$ occurs with probability at least $1 - n^{-c}$ when $n \ge N_3$. Assume $\mathcal{E}_n(c)$ holds. 

Then similar to Theorem \ref{thm:hausdorff_consistency}, we have
\begin{equation}
\label{Rhatp}
    d_H(\mathcal{R}(\widehat{p}), \mathcal{R}(p)) \le \frac{2 \epsilon_\xi^{\log}(n)}{\gamma_{\log}}.
\end{equation}

As shown in the proof of Lemma \ref{lem:automatic_stability}, the adaptive step size $\alpha_n$ satisfies $\inf_{x \in \mathcal{D}} \alpha_n(x) \ge \frac{C_K h^2}{2}$. Similar to Theorem \ref{thm:empirical_scms_convergence} and Corollary \ref{cor:algorithmic_hausdorff}, we can show that
\begin{equation}
\label{hatXm}
d_H(\widehat{\mathcal{X}}_m^{\log}, \mathcal{R}(\widehat{p})) \le \frac{8\epsilon}{\gamma_{\log}} \rho_n^m,
\end{equation}
where $\rho_n = 1 - \frac{C_K \gamma_{\log}}{16} h^2 $

By the triangle inequality for the Hausdorff metric,
\begin{equation}
    d_H(\widehat{\mathcal{X}}_m^{\log}, \mathcal{R}(p)) \le d_H(\widehat{\mathcal{X}}_m^{\log}, \mathcal{R}(\widehat{p})) + d_H(\mathcal{R}(\widehat{p}), \mathcal{R}(p)).
\end{equation}
Substituting \eqref{Rhatp} and \eqref{hatXm} into this inequality yields the total error bound.
\end{myproof}

\end{document}